\documentclass{article}

\usepackage{PRIMEarxiv}

\usepackage[utf8]{inputenc} % allow utf-8 input
\usepackage[T1]{fontenc}    % use 8-bit T1 fonts
\usepackage{hyperref}       % hyperlinks
\usepackage{url}            % simple URL typesetting
\usepackage{booktabs}       % professional-quality tables
\usepackage{amsfonts}       % blackboard math symbols
\usepackage{nicefrac}       % compact symbols for 1/2, etc.
\usepackage{microtype}      % microtypography
\usepackage{lipsum}
\usepackage{amssymb}
\usepackage{fancyhdr}       % header
\usepackage{graphicx}       % graphics

\usepackage{subcaption} % For subfigures
\usepackage{algorithm}%
\usepackage{algorithmicx}%
\usepackage{algpseudocode}%
\usepackage{listings}%
\usepackage{todonotes}
\usepackage{amsmath}

\usepackage[numbers]{natbib} % numeric citations
\usepackage{tcolorbox}
\usepackage{wrapfig}  % For wrapping text around figures/tables
\usepackage{caption}    % Needed for \captionof

\usepackage{setspace}

\title{Hyperdimensional Cross-Modal Alignment of Frozen Language and Image Models for Efficient Image Captioning 
}

\author{
  Abhishek Dalvi \;\;\;\; Vasant Honavar \\ 
  \\
  Artificial Intelligence Research Laboratory\\
  The Pennsylvania State University \\
  \texttt{abd5811@psu.edu \; vuh14@psu.edu} \\
}

\begin{document}
\maketitle

\begin{abstract}

Large unimodal foundation models for vision and language encode rich semantic structures, yet aligning them typically requires computationally intensive multimodal fine-tuning. Such approaches depend on large-scale parameter updates, are resource-intensive, and can perturb pretrained representations. Emerging evidence suggests, however, that independently trained foundation models may already exhibit latent semantic compatibility, reflecting shared structures in the data they model. This raises a fundamental question: can cross-modal alignment be achieved without modifying the models themselves? Here we introduce \textbf{HDFLIM (HyperDimensional computing with Frozen Language and Image Models)}, a framework that establishes cross-modal mappings while keeping pretrained vision and language models fully frozen. HDFLIM projects unimodal embeddings into a shared hyperdimensional space and leverages lightweight symbolic operations -- binding, bundling, and similarity-based retrieval -- to construct associative cross-modal representations in a single pass over the data. Caption generation emerges from high-dimensional memory retrieval rather than iterative gradient-based optimization. We show that HDFLIM achieves performance comparable to end-to-end vision–language training methods and produces captions that are more semantically grounded than zero-shot baselines. By decoupling alignment from parameter tuning, our results suggest that semantic mapping across foundation models can be realized through symbolic operations on hyperdimensional encodings of the respective embeddings. More broadly, this work points toward an alternative paradigm for foundation model alignment in which frozen models are integrated through structured representational mappings rather than through large-scale retraining. The codebase for our implementation can be found at \href{https://github.com/Abhishek-Dalvi410/HDFLIM}{https://github.com/Abhishek-Dalvi410/HDFLIM}
\end{abstract}

% keywords can be removed
\keywords{Hyperdimensional Computing \and Image Captioning \and Frozen Models \and Symbolic Mapping}

\section{Introduction}\label{sec1}

Vision-language models have advanced rapidly with large scale architectures that align visual representations with natural language. By incorporating generative objectives and tighter cross-modal integration \cite{zhu2024minigpt4, LLAVA_VL, yu2022coca, chen2023pali}, these models achieve strong performance on benchmarks such as image captioning \cite{lin2014microsoft_COCO, young_Flickr30k, agrawal2019nocaps} and visual question answering \cite{Antol_VQA, balanced_binary_vqa, balanced_vqa_v2}. Recent approaches either train vision and language components end-to-end for richer multimodal reasoning \cite{wang2024qwen2vlenhancingvisionlanguagemodels, LLAVA_VL}, or adopt modular designs that connect pretrained components for greater efficiency \cite{blip2_sales, Barraco2022TheUE_unreason_CLIPfeat}. Despite these advances, significant limitations remain. End-to-end models are computationally expensive, while modular approaches often sacrifice on adaptability but still require substantial training and can be unstable under continued adaptation, limiting scalability in practice.

Recent work points to an emergent alignment between the internal representations learned by large-scale vision and language models \cite{maniparambil2024vision, platonic-hypothesis, indra_representation}. Remarkably, even when trained in isolation, unimodal models can exhibit substantial semantic compatibility, suggesting that both modalities may converge toward shared latent structure in their representations of the world. This convergence raises the possibility that such models are implicitly recovering elements of a common conceptual substrate. Building on this perspective, this work explores the integration of symbolic reasoning and structured knowledge as a pathway toward vision–language systems that are more interpretable, robust, and data efficient, and that may ultimately support more compositional and generalizable forms of reasoning.

In parallel, Hyperdimensional (HD) computing \cite{Kanerva1988SparseDM, HD_computing_intro_kanerva, gaylor_VSA, plate_HRR, neubert2019introduction} has gained attention as an alternative computational framework. By representing information using extremly high-dimensional bipolar vectors and manipulating them with simple operations such as binding and bundling, HD computing supports efficient, noise-tolerant computation with relatively low overhead. These properties make HD computing well suited to incremental and continual learning, and therefore a natural candidate for exploiting emergent semantic alignment while enabling lightweight symbolic structure.

\textbf{Key Contribution:} Motivated by recent observations between vision and language models, we introduce \textbf{HDFLIM}, a model that leverages pretrained vision and language encoders in a frozen state while harnessing the efficiency and robustness of HD operations. HDFLIM maps features from both modalities into HD space and exploits their shared structure to support symbolic binding, enabling effective learning for image captioning in a single pass through the training data without iterative weight updates or backpropagation over multiple epochs typical of deep learning. The implementation also incorporates partial on-disk learning and optimization at inference, contributing to faster token generation compared to zero-shot, train-free approaches. By integrating emergent alignment with HD computing symbolic operations, HDFLIM provides a scalable, interpretable, and efficient framework that combines the strengths of pretrained models with fast, disk-assisted computation. Our experiments show that HDFLIM produces captions that are semantically richer and more relevant than those from zero-shot, training free methods, while matching the performance of end-to-end vision--language models on reference free and semi-reference free metrics. More broadly, this work points toward an alternative paradigm for foundation model alignment in which frozen models are integrated through structured representational mappings rather than through large-scale retraining.

\section{Related Work}

\subsection{Vision to Language Models}

Vision–language research has advanced rapidly with large-scale models that align visual and linguistic representations. Early approaches such as CLIP \citep{CLIP-radford2021learning} relied on contrastive learning to enable strong zero-shot recognition, but did not support open-ended generation or reasoning. Subsequent models, including CoCa \citep{yu2022coca}, incorporated generative objectives to improve captioning and alignment, though they require computationally intensive end-to-end training\footnote{https://laion.ai/blog/coca/}.

Current methods largely follow two paradigms. The first is end-to-end multimodal training, where vision encoders and language decoders are jointly optimized (e.g., Qwen-VL \cite{wang2024qwen2vlenhancingvisionlanguagemodels}, PaLI \cite{chen2023pali}, Gemini \cite{geminiteam2025geminifamilyhighlycapable}), enabling rich cross-modal reasoning at the cost of substantial computational overhead. The second paradigm adopts modular designs that connect frozen pretrained vision and language models via adapters, such as BLIP-2 \cite{blip2_sales} with the Q-Former. While this improves efficiency, these models often fine-tune the vision encoder and adapter during benchmarking, increasing training cost and risking catastrophic forgetting \cite{FRENCH1999128_Catastrophic, Kirkpatrick_2017_catastrophic} in the vision backbone, which undermines long-term stability and reusability. Related approaches that project visual features directly into the language model’s embedding space (e.g., LLaVA \cite{LLAVA_VL}, MiniGPT-4 \cite{zhu2024minigpt4}) achieve strong reasoning performance but still rely on substantial backpropagation and remain susceptible to forgetting.

Training-free captioning methods such as ConZIC \cite{Zeng_2023_CVPR_conzic} and ZeroCap \cite{Tewel_2022_CVPR_zerocap} shift the burden of learning to inference-time computation. They rely on iterative sampling guided by CLIP similarity or gradient-based steering of an LLM using a CLIP derived loss, thereby avoiding parameter updates. However, this reliance on test-time optimization makes them prone to hallucinations; an issue also noted by \citet{Zeng_2024_CVPR_MeaCap}, and leads to slow inference that scales poorly with caption length and decoder vocabulary size.

In sum, existing paradigms face a trade-off: computational efficiency vs. flexibility, performance, and stability. While modular methods reduce cost, they sacrifice reasoning depth; end-to-end models deliver strong performance at high cost; and hybrid approaches still require extensive training and careful management. These limitations underscore the need for novel, low-cost, incremental architectures that support continuous learning without catastrophic forgetting.

\subsection{Emergent Alignment between Vision and Language Models}

Recent studies have revealed a compelling insight: well-trained unimodal vision and language encoders, when pretrained on large scale datasets, exhibit a remarkable degree of semantic alignment even when trained independently. This suggests that both modalities naturally learn to represent the same underlying physical world, leading to structurally and semantically similar internal representations. 

Building on this intuition for learning using world models, \citet{merullo2023linear-ImagetoText-mapping} demonstrated that a simple linear mapping is sufficient to translate visual representations into language representations for downstream vision–language tasks. Recent models such as VL-JEPA \cite{chen2025vl-jepa}, based on the JEPA \cite{Assran_2023_CVPR_JEPA} architecture, adopt performing prediction directly in latent space rather than generating discrete tokens. VL-JEPA jointly trains vision and language encoders, typically initialized from strong pretrained unimodal models  \cite{assran2025vjepa2selfsupervisedvideo, vera2025embeddinggemma} and aligns them through a joint-predictive objective in latent space. Notably, in a strictly controlled comparison with matched training budgets, VL-JEPA outperforms token-generative VLMs while using fewer trainable parameters, demonstrating higher sample efficiency. Although, VL-JEPA does not explicitly cite cross-modal semantic alignment as a motivating factor and emphasizing instead the general advantages of latent-space prediction; it is plausible that such alignment might play a crucial contributing role. When vision and language representations are already semantically compatible, prediction in latent space becomes both easier and more efficient, enabling more stable learning and reduced optimization complexity compared to token-level generation.

More direct semantic alignment is seen in \citet{maniparambil2024vision} work which demonstrate this alignment by introducing a novel kernel based method that quantifies semantic similarity between the representation spaces of vision and language encoders without cross-modal training. Further supporting this idea, \citet{platonic-hypothesis} provide more explicit evidence using mutual nearest neighbors (MNN) analysis, which reveals strong correspondence between the embeddings of large vision models like vision encoders (e.g., DINOv2 \cite{oquab2023dinov2}) and LLMs (e.g., Llama \cite{touvron2023llamaopenefficientfoundation-llama, grattafiori2024llama3}). These findings suggest that the alignment is not coincidental but rather a consequence of both models learning to capture fundamental, world-grounded concepts. Building on this, the authors propose a compelling hypothesis: \textit{the stronger the unimodal models--both in terms of model capacity and training data--the more aligned their representations become}. Similar findings are discussed by \citet{indra_representation}, who likewise argue that increasingly capable unimodal systems converge toward a shared representational structure. In other words, as unimodal vision and language  models grow more powerful, their representations converge toward a shared, semantically rich space that reflects universal, real-world knowledge.

% \subsection{Linking Semantics: From Emergent Alignment to Symbolic Grounding}

% As vision and language models increasingly align in their representational spaces, neurosymbolic AI emerges as a pivotal framework for robust, interpretable semantic linking (For example usiing knowlegde graphs \cite{delong2024neurosymbolic}). It has been evident in the past years that symbolic systems encompass diverse approaches including differentiable logic programming~\cite{yang2022differentiable}, Logic Tensor Networks~\cite{serafini2016logic}, Neural Theorem Provers~\cite{rocktaschel2017end}. Systems like these are AI enabaled to extract, structure, and reason over knowledge from raw data, generate novel concepts through logical rule induction, and guide neural inference with formal constraints.

% This synergy enables AI systems to ground, map, and reason over shared conceptual structures across modalities. The emergent semantic alignment of vision and language models creates a natural substrate for symbolic binding operations—whether through knowledge graph entity linking, logical rule application, or ontological concept mapping—that enhance downstream task performance while preserving interpretability and robustness~\cite{delong2024neurosymbolic, yang2022differentiable}.

\subsection{Hyperdimensional Computing}

Hyperdimensional (HD) computing \citep{Kanerva1988SparseDM, HD_computing_intro_kanerva, neubert2019introduction, plate_HRR, gaylor_VSA} offers an alternative computational framework in which information is represented using extremely high-dimensional binary or bipolar vectors; typically with dimensionality on the order of $10^4$ or higher—and manipulated through simple algebraic operations. Unlike deep learning approaches that rely on iterative optimization of a loss function, HD computing maps data into a high-dimensional representational space in a largely non-iterative manner, enabling efficient learning, robustness to noise, and straightforward incremental updates.

Following \citet{HD_computing_intro_kanerva, Kanerva1988SparseDM} HD computing framework, we employ either transformed/hashed or randomly generated bipolar hyperdimensional vectors, denoted as $\{-1,1\}^{\beta}$ with $\beta = 50{,}000$. There are also other works, such as \citet{plate_HRR} which work in the continuous space. 

Sets of such hypervectors are composed using hyperdimensional operations to produce multi-bit representations for downstream inference which capture some meaniful information. Similarity between such semantic meaningful hypervectors is measured using a distance metric $d_{\mathcal{H}}$: for bipolar vectors this is typically the normalized Hamming distance. A key property of randomly generated bipolar hypervectors is their near-orthogonality: with high probability, the similarity between any two such vectors $\mathbf{a}$ and $\mathbf{b}$ satisfies $d_{\mathcal{H}}(\mathbf{a}, \mathbf{b}) \approx 0.5$. This property, together with the distributed nature of hyperdimensional representations, yields inherent robustness to noise and supports reliable content-addressable retrieval and associative memory~\citep{Kanerva1988SparseDM,neubert2019introduction}.

% From a representational geometry perspective, the effectiveness of HDC can be understood through Cover’s theorem on the separability of patterns, which states that a nonlinear mapping into a sufficiently high-dimensional space increases the likelihood that patterns become linearly separable. Hyperdimensional representations instantiate this principle explicitly: data are projected into a very high-dimensional space using fixed or structured mappings, after which inference reduces to similarity computation or simple linear decision rules. Unlike kernel methods or deep nonlinear transformations, this expansion is achieved through algebraic composition rather than learned nonlinearities, avoiding iterative optimization and associated instability.

In this work we rely on two fundamental operations:
\begin{itemize}
    \item \textbf{Binding} $\otimes$: a dimension-wise multiplication operation that associates two hypervectors to produce a composite representation that is dissimilar to each constituent. For randomly sampled bipolar hypervectors $\mathbf{a}$ and $\mathbf{b}$, the bound hypervector $\mathbf{c} = \mathbf{a} \otimes \mathbf{b}$ satisfies $d_{\mathcal{H}}(\mathbf{c}, \mathbf{a}) \approx 0.5$ and $d_{\mathcal{H}}(\mathbf{c}, \mathbf{b}) \approx 0.5$, indicating near orthogonality. Binding is commonly used to represent variable-value associations \cite{Kanerva1988SparseDM, Gayler1998MultiplicativeBR_rotate} or even to bind symbolic representations, as explored in prior work by \citet{smolensky1990tensorbinding}.
    \item \textbf{Bundling} $\oplus$: dimension-wise majority operation that aggregates multiple hypervectors into a single hypervector. For randomly sampled hypervectors, the resulting bundled hypervector remains similar to its constituent members while remaining dissimilar to non-members, due to the properties of HD representations \cite{Kanerva1988SparseDM, neubert2019introduction}. Bundling is associative and commutative, and is commonly used to represent sets, multisets, or collections of bound structures.
\end{itemize}

In addition to these operations, \textbf{Rotation/Permutation} is another fundamental operation introduced by \citet{Gayler1998MultiplicativeBR_rotate}, which increases the expressive richness of HD computing. Although, rotation is not used in this work, it represents a promising direction for future extensions.

\emph{Overall, a comprehensive primers on HD computing in context to this work can be found in \citet{HD_computing_intro_kanerva} and \citet{neubert2019introduction}, which introduce its core principles and operations. A theoretical overview of why HD computing is effective grounded in HD statistical properties is provided by \citet{thomas2021theoretical_HD}.}

\paragraph{Illustrative Example: Learning and Inferring an Image Scene}

Consider an image depicting the textual phrase \texttt{"New car on road"}, composed of four words/tokens: $\{w_1, w_2, w_3, w_4\}$. A deep learning based image encoder extracts spatial patches from the image, generating rich semantic representations; similar to the approach used in Vision Transformers \cite{dosovitskiy2020image_VIT}. These patch-level features are subsequently mapped into a HD space (eg. $\approx 50k$ dimension space), preserving their semantic structure; for instance, via Locality Sensitive Hashing (LSH) \cite{IndykLSH, LSHcharikar}.

% We assume access to rich semantic representations of this image via a deep learning-based image encoder, which are further transformed into a HD space such that their semantics isn't lost something using Locality Sensitive Hashing (LSH) \cite{IndykLSH, LSHcharikar}. In this representation, spatially meaningful features are encoded as distinct hypervectors, enabling structured and interpretable processing. 

For simplicity, assume this image HD representations are obtained into three patches/spatial regions: left, center, and right. Each region is associated with a semantic HD vector: $\mathbf{v}_{\text{L,scene}}$, $\mathbf{v}_{\text{C,scene}}$, and $\mathbf{v}_{\text{R,scene}}$, capturing the semantic content within that region (e.g., road on the left, the car in the center, background on the right).

To compactly and symbolically represent the full scene, we introduce three random, role-specific reference hypervectors: $\mathbf{r}_{\text{left}}$, $\mathbf{r}_{\text{center}}$, and $\mathbf{r}_{\text{right}}$. These serve as spatial anchors and are bound to their respective region specific latent scene vectors. These representations are combined scene-level hypervector:

\begin{equation}\label{eq:example_img_encode}
    \mathbf{S} = 
(\mathbf{r}_{\text{left}} \otimes \mathbf{v}_{\text{L,scene}}) \oplus
(\mathbf{r}_{\text{center}} \otimes \mathbf{v}_{\text{C,scene}}) \oplus
(\mathbf{r}_{\text{right}} \otimes \mathbf{v}_{\text{R,scene}})
\end{equation}

Here, $\otimes$ denotes the Binding operation , and $\oplus$ denotes Bundling.

Now, suppose we also have a semantic text encoder; such as a LLM that processes a partial string. In this case, it encodes the substring \texttt{"New car on"}, corresponding to $w_1, w_2, w_3$. Assuming these rich textual features are projected into a HD space (also using LSH but not semantically similar to Visual HD space), we obtain a HD representation $\mathbf{v}_{w_1w_2w_3}$.

To capture the semantic relationship between the visual scene and its textual description, we bind the scene representation $\mathbf{S}$ with the text representation $\mathbf{v}_{w_1w_2w_3}$:

$$
\mathbf{S} \otimes \mathbf{v}_{w_1w_2w_3}
$$

which encodes the joint visual-textual meaning of \texttt{"New car on"} in HD space.

Now, consider a dataset of $N$ such scene-text pairs. We can learn a \textit{prototype representation} for the semantic concept of a car being on \texttt{road} by aggregating the binding of scene and text representations across all relevant examples:

$$
\mathbf{v}_{\text{road}} = \bigoplus_{i=1}^{N} \left( \mathbf{S}^{i} \otimes \mathbf{v}_{w_1w_2w_3}^{i} \right)
$$

Similarly, a separate prototype $\mathbf{v}_{\text{snow}}$ can be constructed for the scene-text pair \texttt{"New car on snow"}, using another set of scene-text pairs.

At test time, the goal is to determine whether an image depicts a new car on \{\texttt{snow}\} or \{\texttt{road}\}. Consider an unseen test image $k$, representing the phrase \texttt{"New car on \{road\}"} (not known since this is a test image). This image is processed using the same encoding pipeline as in training i.e Eq. \eqref{eq:example_img_encode}, yielding a test scene representation $\mathbf{S}^k$.  We can then compute the joint representation with the partial substring \texttt{"New car on"} as:

$$
\mathbf{S}^k \otimes \mathbf{v}_{w_1w_2w_3}
$$

Due to the properties of HD computing ;we can compare this test representation to the learned prototypes using hamming distances (see  \citet{Joshi2016Languageclass_kanerva} an example of classification):

\begin{equation}\label{eq:hamming_compare_example}
    d_{\mathcal{H}}\left( \mathbf{v}_{\text{road}}, \left( \mathbf{S}^k \otimes \mathbf{v}_{w_1w_2w_3} \right)\right) < d_{\mathcal{H}}\left( \mathbf{v}_{\text{snow}},\left( \mathbf{S}^k \otimes \mathbf{v}_{w_1w_2w_3} \right)\right)
\end{equation}

This inequality indicates that the test image is more semantically similar to and indicative of  \texttt{"New car on road"} prototype than to \texttt{"New car on snow"}, enabling accurate prediction based of partial text and image; resembling the task of image captioning. 

It is worth noting that if the substring \texttt{"Latest car on"} ($w_{\psi}w_2w_3$); a string never seen during prototype construction as $w_\psi=\text{\texttt{"Latest"} }$ was never seen, were encoded using the same methodology (i.e., LLM features mapped to HD space) yielding $\mathbf{v}_{w_{\psi}w_2w_3}$. If this HD vector is \textbf{Binded} $\otimes$ with the test image vector $\mathbf{S}^k$ (depicting \texttt{"New car on road"}), a similar result to Eq. \ref{eq:hamming_compare_example} is highly plausible, specifically:
$$
    d_{\mathcal{H}}\left( \mathbf{v}_{\text{road}}, \left( \mathbf{S}^k \otimes \mathbf{v}_{w_{\psi}w_2w_3} \right)\right) < d_{\mathcal{H}}\left( \mathbf{v}_{\text{snow}},\left( \mathbf{S}^k \otimes \mathbf{v}_{w_{\psi}w_2w_3} \right)\right)
$$

since
$$
\mathbf{v}_{\text{\texttt{"New car on"}}} \approx \mathbf{v}_{\text{\texttt{"Latest car on"}}},
$$
The above approximation holds due to the semantic similarity between \texttt{"new"} and \texttt{"latest"}; both convey the notion of recentness; which is captured by the LLM representation and consequently, their corresponding HD representations.

\section{HDFLIM: HyperDimensional computing with Frozen Language and Image Models}

We systematically extend and scale the illustrative example by constructing HD prototypes for each token in a predefined vocabulary $\mathcal{V}$, across all token positions (upto $\ell_{max}$ number of tokens). We refer to this approach as \textbf{HDFLIM}, which leverages SOTA frozen vision and language models without any fine-tuning or gradient-based learning. By establishing a symbolic bridge between modalities through HD computing, \textbf{HDFLIM \emph{enables learning over the entire dataset in a single pass}}; enabling efficient, backpropagation free learning. Crucially, since the foundation vision and language models are used exclusively in inference mode and never updated, the approach avoids catastrophic forgetting while preserving the rich, unimodal semantics in the frozen models.

The following subsections provide a detailed exposition of the HDFLIM framework. For a high-level overview, we summarize the learning, inference, and sampling procedures in Algorithms~\ref{algo:HDFLIM_Learn}, ~\ref{algo:HDFLIM_infer}, and ~\ref{algo:HDFLIM_sample} in the Appendix, respectively, with key steps abstracted to highlight the core workflow.

\begin{figure}[ht!]
    \centering
    \includegraphics[width=0.8\linewidth]{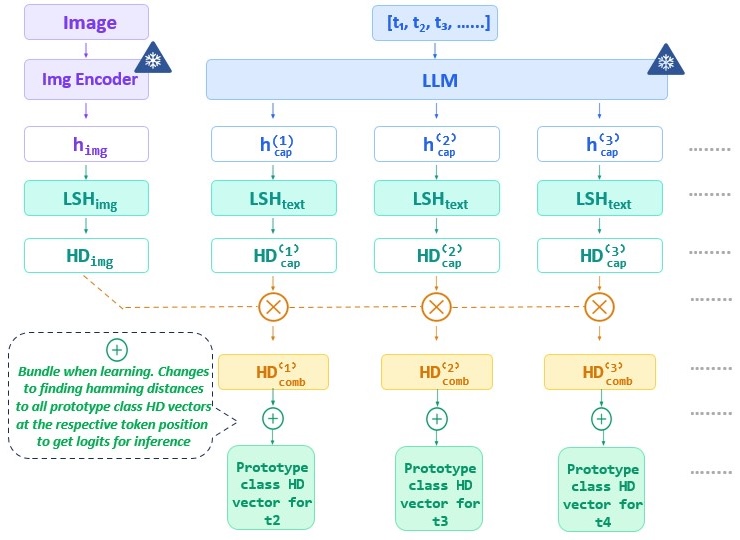}
    \caption{Overview of the HDFLIM algorithm where Image Encoder and LLM are kept frozen. HD representations of the $i^{th}$ token are binded with the HD representation of the image, and this acts as an contextual cue for the next token during the inference stage. }
    \label{fig:HFLIM_overview}
\end{figure}
\subsection{Learning}

The HDFLIM learning process consists of three main steps:

\begin{enumerate}
    \item Extracting visual patch features from the image using a frozen vision model and projecting them into a single HD image vector.
    \item Encoding the caption using a frozen language model and projecting each token’s hidden representation into a corresponding HD token vector.
    \item Binding the image HD vector with the token HD vector and accumulating the result into a position and token specific prototype hypervector.
\end{enumerate}

Learning proceeds by iterating over image-caption pairs only once and progressively accumulating symbolic associations in hyperdimensional space.

\subsubsection{Converting real valued Features to HD vectors using LSH}

A crucial step in the subsequent sections is converting real-valued informative features obtained from the LLM and vision model into an HD representation. To achieve this, we employ Locality Sensitive Hashing (LSH) \cite{IndykLSH, LSHcharikar, Vershyninhyperplane_tessla, Dirksensharp_randomHyperPlane, Oymak2015NearOptimalBF, Dirksen2018NonGaussianHyperplane, Andoni2013BeyondLH}. LSH possesses properties that align with those of HD computing; namely, it is distributional and preserves spatial locality, such that points that are close in the original feature space are likely to produce similar hash codes, as theoretically established in \citet{LSHcharikar, Vershyninhyperplane_tessla}. Moreover, its effectiveness has been empirically validated in various machine learning applications via HD computing \cite{dalvi-causalmatching, Dalvi_HDGL, Dalvi_CHDnet}, further supporting its suitability for this task.

In this work, we use the angular version of LSH proposed by \citet{LSHcharikar}. We first map features from $\mathbb{R}^{d}$ to a HD binary space $\{-1,1\}^{\beta}$, where $\beta \gg d$, using random hyperplane projections. Specifically, we sample $\beta$ random hyperplanes in $\mathbb{R}^{d}$ and assign each coordinate of the HD representation based on whether a feature vector lies on the positive or negative side of the corresponding hyperplane. This yields a $\beta$-dimensional binary sketch for each data point. 

Formally, let $\mathbf{Q} \in \mathbb{R}^{\beta \times d}$ be a matrix whose rows $\mathbf{q}_{1}^{T}, \ldots, \mathbf{q}_{\beta}^{T}$ are independently drawn from $\mathcal{N}(0, I_{d})$. The binary sketch for a $\mathbf{x} \in \mathbb{R}^{d}$ is then given by 
\begin{equation} \label{eq:angular_LSH}
    \mathbf{r} = \text{LSH}_{\mathbf{Q}}(\mathbf{x}) =\mathrm{sign}(\mathbf{Q}\mathbf{x})
\end{equation}

The resulting vector $\mathbf{r}$ is a $\beta$-dimensional binary HD representation, where similar data points produce similar binary sketches. In our work, we set $\beta=50,000$.

In our setup, we apply the LSH mapping to image features extracted from a frozen vision model and text features from a frozen large language model (LLM). To ensure consistency across training and inference, we use two random projection matrices: $\mathbf{Q}_{\mathcal{I}} \in \mathbb{R}^{\beta \times d_{\mathcal{I}}}$ for image features and $\mathbf{Q}_{\mathcal{C}} \in \mathbb{R}^{\beta \times d_{\mathcal{C}}}$ for caption features, where $d_{\mathcal{I}}$ and $d_{\mathcal{C}}$ denote the respective feature dimensions. To concisely represent the LSH transformation defined in Eq. \eqref{eq:angular_LSH}, we introduce the notation 
$$
\text{LSH}_{\mathbf{Q}_{\mathcal{I}}}(\cdot): \mathbb{R}^{d_{\mathcal{I}}} \to \{-1, 1\}^{\beta};\quad \text{LSH}_{\mathbf{Q}_{\mathcal{C}}}(\cdot): \mathbb{R}^{d_{\mathcal{C}}} \to \{-1, 1\}^{\beta}
$$ 
which map input features to their corresponding HD binary sketches in the high. These functions will be used throughout the paper for clarity and notational convenience.

Note that, both $\mathbf{Q}_{\mathcal{I}}$ and $\mathbf{Q}_{\mathcal{C}}$ are initialized only once at the beginning of the learning phase, saved to disk, and reused consistently during both training and inference. The symbolic prototypes learned during training are constructed based on these LSH projections, and therefore the same matrices are used during inference.

\subsubsection{Computing the Image HD Vector}\label{sec:compute_imgHD_vec}

We restrict ourself to images of resolution $512 \times 512$ in this work. This ensures that each image yields the same number of patch features, resulting in a consistent number of patches across all inputs.

Specifically, let $I$ denote the input image. We first pass $I$ through a frozen vision encoder, $\text{Encoder}_{\text{VM}}(I)$, to obtain patch level hidden features $\mathbf{h}_{\text{img}} \gets \text{Encoder}_{\text{VM}}(I)$, which have shape $(n_p, d_I)$, where $n_p$ is the number of patches and $d_I$ is the dimensionality of the patch representations.

We then apply angular LSH \cite{LSHcharikar} to obtain HD representation of the patch features. Each patch feature $\mathbf{z}_j \in \mathbb{R}^{d_I}$ (for $j = 1, \dots, n_p$) is mapped to a binary vector $\mathbf{r}_i \in \{ -1, +1 \}^\beta$ via the transformation:
$$
\mathbf{r}_j = \text{LSH}_{\mathbf{Q}_{\mathcal{I}}}(\mathbf{z}_j); \quad \forall j \in \{1, \dots, n_p\}
$$
 The resulting binary vectors $\{\mathbf{r}_j\}_{j=1}^{n_p}$ serve as the binary HD representations of the individual patches.

To encode positional or spatial information of the patches, we seed random bipolar vectors $\mathbf{s}_j \in \{ -1, +1 \}^\beta$ for each patch $j \in \{1, \dots, n_p\}$. These vectors act as positional variables, preserving spatial context during aggregation. It is much like storing information in a database row, like \citet{HD_computing_intro_kanerva} has shown in previous work.

We then perform Binding between the bipolar patch representation $\mathbf{r}_j$ and the positional code $\mathbf{s}_j$, followed by Bundling across all patches. The final image level HD representation is constructed as:
$$
\mathbf{HD}_{\text{img}} =  (\mathbf{s}_1 \otimes \mathbf{r}_1) \oplus  (\mathbf{s}_2 \otimes \mathbf{r}_2) \oplus \dots \oplus (\mathbf{s}_{n_p} \otimes \mathbf{r}_{n_p}).
$$
We compactly represent the above LSH transformation followed by patch information aggregation via bundling and binding using  $\mathbf{HD}_{\text{img}} \gets \text{transform}^{\text{HD}}_{\text{img}}(\mathbf{h}_{\text{img}})$ . This construction ensures that $\mathbf{HD}_{\text{img}}$ encodes both rich semantic content from the visual features and structured positional information. Note that, the random seeding of $\mathbf{Q}_{\mathcal{I}}$ and $\mathbf{s}_i; \forall i \in \{1, \dots, n_p\}$ is done only once at the start of the algorithm. These random vectors are stored throughout the learning and then further also used for inference since learning is done using these initializations.

\subsubsection{Computing the Token HD Vector}

Given a caption $C$ with $n_c$ tokens, we process it sequentially, considering the subsequence $t_1, \ldots, t_i$ at each step $i \in [1, n_c-1]$. The first token $t_1$ is always a fixed prefix (e.g., a start token or a CLIP-style prefix such as ``This image shows'') across all captions. At each step $i$, the subsequence $t_1, \ldots, t_i$ is passed through a LLM, which is as an autoregressive causal decoder. We extract the last hidden representation i.e., the hidden state immediately preceding the projection into the vocabulary space and denote this as:
$
\mathbf{h}_{\text{cap}}^{(i)} \gets \text{Encoder}_{\text{LLM}}(t_1, \ldots, t_i),
$
which has shape $(i, d_C)$, where $d_C$ is the dimensionality of the hidden state.

Due to the causal masked self-attention mechanism in the LLM, each token at position $k$ only encodes information from its preceding tokens $t_1, \ldots, t_{k-1}$. This autoregressive property ensures that $\mathbf{h}_{\text{cap}}^{(i)}$ captures the cumulative semantic context up to token $t_i$. To convert this dense vector to HD space we use LSH:

$$
\mathbf{HD}_{\text{cap}}^{(i)} = \text{LSH}_{\mathbf{Q}_{\mathcal{C}}}(\mathbf{h}_{\text{cap}}^{(i)})
$$

We again denote this transformation as:
$
\mathbf{HD}_{\text{cap}}^{(i)} \gets \text{transform}^{\text{HD}}_{\text{cap}}\left( \mathbf{h}_{\text{cap}}^{(i)} \right)
$ for ease of explanation in the pseudo-code.

\subsubsection{Binding and Prototype Accumulation}

The image and token-level HD representations are combined using a binding operation, which produces a composite hypervector that encodes joint visual-linguistic context. Specifically, we bind the image hypervector $\mathbf{HD}_{\text{img}}$ with the caption hypervector up to token $i$, $\mathbf{HD}_{\text{cap}}^{(i)}$:
$$
\mathbf{HD}_{\text{comb}}^{(i)} \gets \mathbf{HD}_{\text{img}} \otimes \mathbf{HD}_{\text{cap}}^{(i)}.
$$

\textit{The resulting hypervector $\mathbf{HD}_{\text{comb}}^{(i)}$ encodes the integrated context of both the image and the caption up to token $t_i$, and serves as a contextual cue for predicting the next token $t_{i+1}$.}

To support efficient token prediction, we maintain a prototype memory $\mathbf{HD}_{\text{pred}}$ of shape $({ \ell_{max} \times |\mathcal{V}| \times \beta})$, where: the first dimension corresponds to the predicted token position $i+1$, the second dimension indexes over the vocabulary $\mathcal{V}$, and the third dimension is the hypervector length $\beta$.

For each step $i$, we accumulate the composite hypervector $\mathbf{HD}_{\text{comb}}^{(i)}$ into the prototype memory corresponding to the next token $t_{i+1}$:
$$
\mathbf{HD}_{\text{pred}}[i+1, t_{i+1}, :] \gets \mathbf{HD}_{\text{pred}}[i+1, t_{i+1}, :] + \mathbf{HD}_{\text{comb}}^{(i)}.
$$

Initially, $\mathbf{HD}_{\text{pred}}$ is initialized to zero. As each caption is processed token by token, the accumulated composite hypervectors $\mathbf{HD}_{\text{comb}}^{(i)}$ are aggregated into the prototype memory based on the corresponding next token $t_{i+1}$. This process gradually builds a distribution of contextual prototypes; each representing the expected visual-linguistic context for a specific token position and vocabulary item which will be used during inference.

Note that after iterating over the full dataset, the \textit{prototype memory is binarized} $\mathbf{HD}_{\text{pred}} \gets \text{sign}(\mathbf{HD}_{\text{pred}})$; and made ready for inference. Note that during inference, the next token is predicted by identifying the prototype in $\mathbf{HD}_{\text{pred}}$ most similar to the current context i.e $\mathbf{HD}_{\text{comb}}^{(i)}$ (which is binary), leveraging the accumulated semantic representations from the learning phase. This process of inference will be more explained in detail in the next section.

\subsection{Inference}
During inference, only the image is available, and the caption is generated autoregressively. The process closely mirrors the learning phase, with one key distinction: the model predicts each token sequentially using the accumulated prototypes stored in $\mathbf{HD}_{\text{pred}}$, which encode learned visual-linguistic semantic context.

We begin with a partial caption containing only the fixed prefix token $\hat{t}_1$, which corresponds to the initial prefix token used during training. Let $i$ denote the index of the last predicted token in the current caption; initially, $i = 1$, as the caption consists of only the prefix token.

At each step $i$, the following steps are performed:

\begin{enumerate}
    \item  Compute the image hypervector $\mathbf{HD}_{\text{img}}$, representing the visual content of the input image in the HD space.
    
    \item  Obtain the HD representation $\mathbf{HD}_{\text{cap}}^{(i)}$ of the current partial caption $\hat{t}_1, \ldots, \hat{t}_i$ by processing each token sequentially through the same HD transformation used during training.
    
    \item Compute the combined visual-linguistic context via binding the hypervecotrs from Step 1 and 2:
    $
    \mathbf{HD}_{\text{comb}}^{(i)} \gets \mathbf{HD}_{\text{img}} \otimes \mathbf{HD}_{\text{cap}}^{(i)}.
    $
    This composite hypervector encodes the integrated context of the image and the partial caption, serving as a dynamic cue for predicting the next token.
\end{enumerate}

To predict the next token $\hat{t}_{i+1}$, we compute logits over the vocabulary using the prototype memory $\mathbf{HD}_{\text{pred}}$. Specifically, we focus on the submatrix $\mathbf{HD}_{\text{pred}}[i+1, :, :]$, which contains all prototypes accumulated during training for the $(i+1)$-th token position.

The logits are derived based on the Hamming distance between $\mathbf{HD}_{\text{comb}}^{(i)}$ and each prototype in $\mathbf{HD}_{\text{pred}}[i+1, :, :]$; computed as follows:
$$
\textbf{Logits}_{\textbf{HD}} = \beta - \mathbf{d}_\mathcal{H}\left(\mathbf{HD}_{\text{pred}}[i+1, :, :], \mathbf{HD}_{\text{comb}}^{(i)}\right),
$$
where $\beta$ is the hypervector dimensionality. This formulation ensures that a smaller Hamming distance (i.e closer to the prototype vocab vector) corresponds to a higher logit, reflecting greater similarity in semantic context.

\paragraph{Logit Mixing}

The next token $\hat{t}_{i+1}$ is typically selected based on the logits produced by a vision-language model. However, relying solely on these logits in \textbf{HDFLIM} often leads to poor linguistic structure, grammatical errors, or semantic inconsistencies; especially as the caption length increases. Since \textbf{HDFLIM} is trained on grammatically correct captions and performs autoregressive generation, any deviation from grammatical or semantic norms at an early step $i$ can propagate and amplify over subsequent steps, with the error becoming increasingly severe as $k$ grows (i.e., at step $i+k$).

To mitigate this issue, we incorporate additional linguistic guidance by fusing the \textbf{HDFLIM} logits with pure LLM logits, which are derived exclusively from the previously generated tokens. Since the LLM has been trained on vast amounts of natural text, its logits encode strong grammatical and semantic priors, leading to more fluent and contextually coherent predictions. To effectively combine these complementary signals, we employ a weighted fusion strategy that dynamically balances the vision-language and language-only components, ensuring that the final token prediction benefits from both visual grounding and linguistic fluency.

This process can also be interpreted as \textit{Logit Deflection}: correcting potentially erroneous or ungrammatical predictions from \textbf{HDFLIM} by steering them toward more linguistically valid alternatives. This is similar to Logit biasing/averaging performed in \citet{wang2025think_logitavg} but for open ended reasoning. Specifically, we perform the following:

\begin{equation}\label{eq:logit_deflection}
    \textbf{Logits}_{\textbf{vocab}} = \frac{\textbf{Logits}_{\textbf{HD}}}{\max\left(\textbf{Logits}_{\textbf{HD}}\right)} + 0.15 \cdot \frac{\textbf{getLogits}_{\textbf{LLM}}(\mathbf{h}_{\text{cap}}^{(i)})}{\max\left(\textbf{getLogits}_{\textbf{LLM}}(\mathbf{h}_{\text{cap}}^{(i)}) \right)}
\end{equation}

Here, $\textbf{getLogits}_{\textbf{LLM}}(\mathbf{h}_{\text{cap}}^{(i)})$ returns the LLM’s next-token probabilities based on the hidden state $\mathbf{h}_{\text{cap}}^{(i)}$ encoding the partial caption history.

We empirically found that a scaling factor of $0.15$ provides a stable and effective balance. A larger value tends to overpower the vision-guided signal, degrading image alignment, while a smaller value fails to sufficiently improve grammatical quality, leading to semantically weak outputs. This optimal trade-off was consistently observed across multiple evaluation settings, underscoring the importance of careful calibration in hybrid language generation.

The next token $\hat{t}_{i+1}$ is selected based on these logits. While a simple argmax strategy (i.e., selecting the token with the highest logit) is possible, we employ a more robust sampling mechanism mentioned in the next sections.

This process is repeated iteratively: the newly predicted token is appended to the caption, and the loop continues until either a stop token (e.g., EOS) is generated or the maximum caption length is reached, at which point caption generation terminates.

\subsection{Frozen Model Choices}

For the frozen vision and language components in HDFLIM, we employ DINOv3\cite{simeoni2025dinov3} with CLIP-style patch features; referred to as \texttt{DINOv3.txt} \cite{Jose_dinov2_txt} as the vision backbone, paired with Qwen3-4B-Base \cite{yang2025qwen3} as the LLM. This combination forms the foundation of our multimodal system.

The selection of DINOv3 is motivated by its emergence as a SOTA, general purpose vision backbone, capable of supporting a wide array of downstream tasks through lightweight, task specific adapters. The \texttt{DINOv3.txt} is one such adapter which produces high-quality, semantically rich patch embeddings that encode CLIP-style visual semantics, making it particularly well suited for tasks such as semantic segmentation. Meanwhile, Qwen3-4B excels in natural language understanding and generation, offering strong reasoning, coherence, and contextual awareness.

Together, these models provide a powerful and modular architecture: both remain fully frozen, preserving their powerful pre-trained capabilities while enable effective multimodal collaboration through HD-FLIM's lightweight symbolic bridge; specifically designed to translate vision based patch representations into language compatible tokens. This interface allows the vision and language models to operate independently and efficiently, while enabling structured, high fidelity interaction across modalities.

\subsection{CLIP-Guided Token Sampling}

Our approach integrates visual guidance into text generation by combining HDFLIM logits with CLIP based vision-language alignment scores at each decoding step. Since our frozen vision model already provides CLIP features, we leverage them to ensure text generation remains aligned with the visual content.

We apply standard sampling techniques: temperature scaling, repetition penalty, and nucleus sampling; to generate candidate tokens. For each candidate, we append it to the current sequence, decode the text, and encode it using CLIP's text encoder. We then compute similarity between the resulting embeddings and the input image embedding $\mathbf{h}_{\text{img}}$ from DINOv3.txt.

The final token selection combines normalized CLIP similarity scores with HD-FLIM logits score using a weighted sum:  
$$ \text{clip\_weight} \times \mathbf{clip\_scores} + (1 - \text{clip\_weight}) \times \mathbf{HD\_scores} $$  
This balances linguistic fluency with visual grounding and the parameter $\text{clip\_weight} \in [0,1]$ controls the trade-off. The overall sampling psuedocode is shown in Algorithm \ref{algo:HDFLIM_sample}

\subsection{Extended positional search over Prototype classes}

As noted, our framework maintains a memory Prototype of vector classes over the vocabulary at different positional contexts. During inference (as described in Algorithm \ref{algo:HDFLIM_infer}, Line 18), the vanilla algorithm computes logits as follows:

$$
\textbf{Logits}_{\textbf{HD}} = \beta - \mathbf{d}_\mathcal{H} (\textbf{HD}_{\textbf{pred}}[i+1,:,:], \mathbf{HD}^{(i)}_{\text{comb}})
$$

This computation is limited to the $i+1$-th token, i.e, only one position in the sequence. While this approach may be acceptable if prototypes are learned over a vast number (scale of billions) of data samples, our learning process is constrained to a relatively small dataset: 13 million image-caption pairs. This limitation can lead to suboptimal performance, particularly in the context of autoregressive language modeling of HD-FLIM.

To illustrate, consider a scenario where two similar sentences differ only in a minor syntactic variation: \\
\texttt{"This image shows the new object..."} vs \texttt{"This image shows new object...."}

If the first sentence was present in the training data, but the second was not, a model trained only on the first may fail to generate the second; despite the two being semantically similar. In such cases, the model may not have sufficient exposure to capture these subtle variations in language patterns.

To address this issue, we propose an improvement: instead of considering only one position (the next token), we \textit{search over a window of nearby prototypes}. Specifically, we define the logits as:

$$
\textbf{Logits}_{\textbf{HD}} = \text{arg max}_{w \in \{1,\dots,W\}} (\beta - \mathbf{d}_\mathcal{H} (\textbf{HD}_{\textbf{pred}}[i+w,:,:], \mathbf{HD}^{(i)}_{\text{comb}}))
$$

Here, $W$ is the maximum window size, representing the number of nearby positional prototypes to consider. The maximum over this window captures the \textit{most responsive prototype} for prediction. This approach allows the model to benefit from neighboring context but comes at a computational overhead (i.e computing hamming distances over multiple slices). Importantly, this strategy is inspired by traditional neural networks, where a maximum activation often guides the model to select the most relevant feature or token during inference.

\section{Implementation Details}

\paragraph{Frozen Vision Model Resolution.}
All experiments use a fixed image resolution of $512 \times 512$. Although the backbone vision encoder supports variable-length patch sequences, our HD conversion pipeline requires a fixed patch count for algorithmic compatibility and efficiency. At $512 \times 512$, the DINOv3.txt model produces 1025 tokens (1 CLS token + 1024 visual patch tokens), and our architecture is designed around this configuration i.e $n_p=1025$ in Section \ref{sec:compute_imgHD_vec}. This resolution also matches common practice in many SOTA vision systems, offering a favorable trade-off between information content and computational cost. We therefore adopt the same setting for inference. Extensions to higher resolutions via sliding windows or patch-wise HDFLIM processing are possible but left for future work.

\paragraph{Partial Ondisk Learning and optimized Inference.}
Training and inference are performed using a single NVIDIA A100-40GB GPU. The system is implemented in Python using NumPy (memmap) and PyTorch for GPU-accelerated computation and HD operations. Due to the large prototype matrix, HDFLIM employs partial on-disk learning, combined with bit-packing for improved speed and memory efficiency during inference. 

\textbf{A detailed description of the on-disk learning procedure, runtime for HDFLIM Learning Phase, as well as the inference optimization with bit-packing, is provided in Appendix \ref{appendix:implementation_details}.}

\section{Experiments}

In this section, we describe the experiments performed for HDFLIM.

\subsection{Datasets for Learning}

For the learning phase of HD-FLIM, we learns two versions of prototype memory: one based on the \texttt{Karpathy Train} split \cite{karpathy2015deep} of the \texttt{COCO} dataset \cite{lin2014microsoft_COCO}, which consists of approximately 82,000 images, each annotated with five captions; typically ranging from 10 to 20 tokens in length.  To increase diversity in the captions, we expand this dataset by duplicating each image five times and assigning one of the five annotated captions to each image. Specifically, we generate approximately 410,000 image-caption pairs, with images repeated but associated with different captions.

The second version is trained on the \texttt{PixelProse} dataset \cite{singla2024pixelprose}. Our downloaded version contains approximately 13 million image-caption pairs, with captions averaging around 100 tokens in length. Since our HDFLIM framework is constrained to a maximum sequence length of 41 tokens, we truncate each caption to the first 41 tokens

It is important to note that no validation data is used in this learning process. The HDFLIM learning is performed solely on these constructed datasets.

\subsection{Datasets for Evaluation}

We evaluate HDFLIM on three standard benchmarks: the \texttt{Karpathy COCO-Test} split \cite{lin2014microsoft_COCO, karpathy2015deep}, a widely used test set from the COCO dataset and the \texttt{NOCAPS Validation} split \cite{agrawal2019nocaps}, a zero-shot captioning benchmark designed to assess generalization to unseen caption styles and content. The \texttt{NOCAPS Validation} split is divided into three domains: In-Domain, Near-Domain, and Out-of-Domain; corresponding to the
similarity of depicted objects to COCO dataset classes.

\subsection{Evaluation Metrics}

For benchmarking, we evaluate the performance of \textbf{HD-FLIM} using the following standard metrics commonly used in image captioning: \textbf{BLEU@4} \cite{BLEU_metric},  \textbf{METEOR (M)} \cite{denkowski-lavie-2014-meteor}, \textbf{CIDEr (C)} \cite{vedantam2015cider}, \textbf{SPICE (S)} \cite{anderson2016spice}. These metrics are typically used in conjunction with human-annotated references. While they are widely accepted in the literature, these metrics have known limitations \cite{hessel2021clipscore, sarto2023positive_PACS, sarto2025positive2_PACS, lee-etal-2024-fleur}; particularly their reliance on n-gram statistics, which can undervalue semantically rich or syntactically correct captions.

In response to these limitations, recent research has introduced more robust metrics such as \textbf{CLIP-S} \cite{hessel2021clipscore}: A reference free metric that evaluates the similarity between generated captions and images and \textbf{RefCLIP-S} \cite{hessel2021clipscore}: An extension of CLIP-S that also weighs the human anotated references using CLIP Text Encoder.

While newer metrics like PAC-S \cite{sarto2023positive_PACS}, PAC-S++ \cite{sarto2025positive2_PACS}, FLEUR \cite{lee-etal-2024-fleur} have been proposed to address further shortcomings of traditional metrics, we focus our evaluations on \textbf{CLIP-S} and \textbf{RefCLIP-S}, as they are widely adopted of the newer metrics in the literature, and have been shown to correlate well with human judgments.
Note that, for all the mentioned evaluation metrics, higher values indicate better performance.

\subsection{Baselines and Comparison Models}

We group the comparison methods according to the level of supervision and training complexity, progressing from train-free approaches to fully supervised models.

\paragraph{Training-Free Methods:}
Our primary baselines are \textbf{ZeroCap}~\cite{Tewel_2022_CVPR_zerocap} and \textbf{ConZIC}~\cite{Zeng_2023_CVPR_conzic}, which represent SOTA training free image captioning approaches. ZeroCap combines CLIP with a text decoder and performs gradient-based optimization at inference time to guide caption generation toward higher CLIP similarity. In contrast, ConZIC employs a non-autoregressive Gibbs-BERT framework that iteratively refines captions using CLIP guidance.

Although our method is not strictly training-free, it learns from data in a single pass through all the data points and avoids repeated backpropagation or iterative optimization. As such, it occupies a middle ground between train free methods and conventional deep learning approaches. ZeroCap and ConZIC therefore serve as key baselines, enabling a fair and meaningful comparison as there are the strongest training-free methods.

\paragraph{Memory-Based Captioning:}
We also compare against \textbf{MeaCap}~\cite{Zeng_2024_CVPR_MeaCap}, which constructs a structured Subject-Predicate-Object textual memory using the CC3M dataset. Caption generation is performed via CLIP-based retrieval from this memory, followed by prompt ensembling using CBART~\cite{he-2021-parallel_CBART} to produce the final caption.

\paragraph{Text-Only Training and Text Memory Models:}
Next, we consider methods trained using text-only supervision, with or without an explicit text memory. These include \textbf{MAGIC}~\cite{su2022language_MAGIC}, \textbf{ViECap}~\cite{fei2023transferable_ViECap}, and \textbf{CapDec}~\cite{nukrai-etal-2022-text_CapDec}.

\textbf{MAGIC} trains only a pretrained language decoder (eg. GPT-2 \cite{radford2019language_gpt2}) on caption text. At inference time, it generates candidate captions using constrained decoding and selects the best caption based on CLIP similarity with the input image. As a result, MAGIC is CLIP-model agnostic and can be plugged into different CLIP variants without retraining.

In contrast, \textbf{ViECap} and \textbf{CapDec} explicitly leverage CLIP embeddings during training. The CLIP text encoder is kept frozen, and a decoder is trained on top of text representations, optionally augmented with a textual memory bank. Because the decoder is learned with respect to a specific CLIP embedding space, these methods are not plug-and-play across different CLIP variants. Owing to the shared embedding space of CLIP, they assume that a decoder trained on text embeddings can generalize to image embeddings at inference time.

\paragraph{End-to-End Image-Text Models:}
Finally, we compare against fully supervised image captioning models trained end-to-end on image-text pairs. This includes two variants of \textbf{CLIP-Captioner}~\cite{Barraco2022TheUE_unreason_CLIPfeat}, using ViT-B/32 and ViT-L/14 CLIP backbones, where a decoder is trained on top of CLIP image features.

To benchmark against large-scale vision--language models, we also include \textbf{Qwen2-VL}~\cite{wang2024qwen2vlenhancingvisionlanguagemodels}, \textbf{7B variant}; which is a SOTA VL model pretrained on web-scale image-text data (scale of $>$1B data points). In addition to the base model (denoted \textbf{Qwen2-VL}$_{\text{Base}}$), we report results from a version fine-tuned on the Karpathy COCO train split using LoRA \cite{hu2022lora} for one epoch (denoted \textbf{Qwen2-VL}$_{\text{FT}}$), following common benchmarking practices for large pretrained VL models (e.g.,~\citet{blip2_sales}) to improve performance on the target benchmarks.

For all of the above methods, if any training or fine-tuning is involved, it is performed on the Karpathy COCO train split, where applicable. \textbf{MeaCap}, \textbf{ViECap}, and \textbf{CapDec} are evaluated only in the zero-shot setting on the nocaps dataset, as they are primarily designed for zero-shot captioning and because some evaluation metrics are unavailable for COCO.

\subsection{Parameters configurations for HDFLIM}

We evaluate HDFLIM trained on two datasets: COCO and PixelProse, denoted as \textbf{HDFLIM (C)} and \textbf{HDFLIM (P)} respectively. Unless otherwise stated, both models use identical sampling parameters: temperature $=1.0$, repetition penalty $=1.1$, top-$k=80$, top-$p=0.95$ (nucleus sampling), and clip-weight $=0.5$.

By default, the window parameter is set to $W=3$. The only exception is the COCO and nocaps evaluations (Tables~\ref{tab:COCO_performance}, \ref{tab:nocaps_performance} and \ref{tab:coco_BART_improvement}), where HDFLIM (C) uses $W=1$, as these datasets are closer (or have domains closer) to the training distribution and do not require additional OOD enhancement. For all other experiments, $W=3$ is used unless explicitly noted.

During inference, generation is capped at 15 tokens with early stopping upon producing a full stop, reflecting the single sentence structure of standard captioning datasets. Because PixelProse contains long-form descriptions, truncation may yield incomplete sentences. We therefore paraphrase the generated fragment using a prompt passed through our frozen language model of HDFLIM to produce a fluent, self-contained caption. The prompt template additionally normalizes overly specific details common in PixelProse into the more generic phrasing (eg. no proper nouns for people) typical of benchmark datasets. The prompt template can be found in the Appendix \ref{appendix:config}

\subsection{Results}

We present the main experimental results in this section. Full details of the main results, additional studies and qualitative examples of our predictions are provided in the Appendix \ref{appendix:extended_results} and \ref{appendix:example_predictions}.

\paragraph{Performance on COCO Dataset:}  
In Table \ref{tab:COCO_performance}, both variants of HDFLIM perform strongly compared to SOTA, including Qwen2VL. Train free methods: ZeroCap and ConZIC (based on CLIP-S and RefCLIP-S) also show competitive results, though they are prone to hallucinations. While ConZIC improves caption quality by boosting CLIP-S scores, it still suffers from hallucinations. Traditional metrics (e.g., CIDER, SPICE) remain low; expected for zero-shot, no-training methods,although notably low SPICE scores indicate excessive irrelevant content.

\begin{table}[h!]
\centering
\caption{Performance on Karpathy Test Split of COCO. Max number of generated tokens for HDFLIM and Qwen2VL are limited to 18. $^{\top}$values were taken from \citet{bianchi2025patchcaptionallunified_patchioner}}
\begin{tabular}{  l  | | c c c c| c c}
\hline
Model
    & B@4 & M & C & S & CLIP-S & RefCLIP-S  \\
    \hline
ZeroCap \cite{Tewel_2022_CVPR_zerocap} & 2.6 & 11.5 & 14.6 & 5.5 & 87.0 & 79.0\\

ConZIC \cite{Zeng_2023_CVPR_conzic} & 1.3 & 11.2 & 13.2 & 5.0 & 99.0 & 83.0\\
\hline
\hline

MAGIC \cite{su2022language_MAGIC} & 12.9 & 17.2 & 48.3 & 10.9 & 74.0 & 77.0\\
ViECap$^{\top}$ \cite{fei2023transferable_ViECap} & 26.7 & 24.0 &  89.6 & 17.5  & 75.6  & -\\
\hline
\hline
CLIP-Captioner-B/32 \cite{Barraco2022TheUE_unreason_CLIPfeat} & 36.0 & 27.8 & 114.9 & 20.8 & 75.0 & 80.3\\
CLIP-Captioner-L/14 \cite{Barraco2022TheUE_unreason_CLIPfeat} & 38.7 & 29.3 & 126.0 & 22.5 & 75.4 & 81.1\\
Qwen2VL$_{\text{Base}}$ \cite{wang2024qwen2vlenhancingvisionlanguagemodels}     & 16.9 & 26.0 & 47.1 & 20.3 & 81.0 & 81.9\\
Qwen2VL$_{\text{FT}}$ \cite{wang2024qwen2vlenhancingvisionlanguagemodels}       & 40.0& 30.7 & 137.5  & 24.2 & 78.6 & 84.0\\
\hline
\hline
HDFLIM (C)     & 6.9 & 16.6 & 49.2 & 12.3 & 75.3 & 79.2 \\
HDFLIM (P)    & 7.8 & 15.2 & 38.9 & 10.4 & 73.6 & 77.8 \\
\hline
\end{tabular}
\label{tab:COCO_performance}
\end{table}

In contrast, our method achieves significantly higher SPICE scores than MAGIC, while maintaining CLIP-S scores on par with CLIP-Captioner, an end-to-end model. Notably, HDFLIM (C) outperforms HDFLIM (P) on COCO, as it is trained on COCO data, highlighting the benefit of domain alignment.

\paragraph{Zero-Shot Performance on NoCaps:}  
On NoCaps dataset, similar trends as before, emerge as seen in Table \ref{tab:nocaps_performance}. Despite being end-to-end trained, CLIP-Captioner fails to generalize well, underperforming HDFLIM in CLIP-S. HDFLIM (P) excels in CLIP-S due to broader training data, but HDFLIM (C) surpasses it in CIDER and SPICE on in-domain and near-domain settings; consistent with its COCO specific training. However, for out-of-domain captions, HDFLIM (C) degrades, while HDFLIM (P) remains more robust.

\begin{table}[h!]
\centering
\caption{Zero-shot performance on NoCaps Val split with CIDEr (C), SPICE (S) and CLIP-Scores metrics. In, Near and Out are Domains w.r.t to the objects/concepts with the MS-COCO dataset. * results were taken from \citet{Zeng_2024_CVPR_MeaCap} and only contained CIDEr score. $^{\dagger}$ values were taken from \citet{fei2023transferable_ViECap}}
\begin{tabular}{  l || cc| cc |cc |cc|c}
\hline
& \multicolumn{8}{|c|}{NoCaps Domains}  &  \\
\hline
Model 
& \multicolumn{2}{|c|}{In} 
& \multicolumn{2}{|c|}{Near} 
& \multicolumn{2}{|c|}{Out} 
& \multicolumn{2}{|c|}{Overall} & Overall\\
\hline
& C & S & C & S & C & S & C & S  &  CLIP-S\\
\hline

ZeroCap* \cite{Tewel_2022_CVPR_zerocap} &
13.3 & - & 14.9 & - & 19.7 & - & 16.6 & - & -  \\
ConZIC* \cite{Zeng_2023_CVPR_conzic} &
13.7 & - & 15.8 & - & 18.3 & - & 16.9 & - & - \\
\hline
\hline
MAGIC$^{\dagger}$ \cite{su2022language_MAGIC}
& - & - & - & -& - & - & - & - & 66.2 \\
CapDec$^{\dagger}$ \cite{nukrai-etal-2022-text_CapDec}
& 60.1 & 10.2 & 50.2 & 9.3 & 28.7 & 6.0 & 45.9 & 8.3 & 69.2 \\
ViECap$^{\dagger}$ \cite{fei2023transferable_ViECap}
& 61.5 & 10.4 & 64.3 & 9.9 & 65.0 & 8.6 & 66.2 & 9.5 & 75.4\\
MeaCap* \cite{Zeng_2024_CVPR_MeaCap} &
35.3 & - & 39.0 & - & 45.1 & - & 40.2 & -  & - \\
\hline
\hline
CLIP-Captioner-B/32 \cite{Barraco2022TheUE_unreason_CLIPfeat}
& 89.6 & 12.7 & 75.5 & 11.8 & 51.2 & 9.6 & 72.6 & 11.5 & 68.9 \\
CLIP-Captioner-L/14 \cite{Barraco2022TheUE_unreason_CLIPfeat}
& 104.5 & 13.5 & 92.2 & 13.0 & 67.6 & 10.7 & 89.0 & 12.6 & 70.8 \\
\hline
\hline
Qwen2VL$_{\text{Base}}$ \cite{wang2024qwen2vlenhancingvisionlanguagemodels}
& 48.5 & 14.8 & 51.0 & 14.5 & 57.4 & 14.7 & 53.3 & 14.6 & 81.4 \\
Qwen2VL$_{\text{FT}}$  \cite{wang2024qwen2vlenhancingvisionlanguagemodels}
& 118.4 & 15.3 & 120.0 & 15.6 & 123.1 & 15.7 & 122.3 & 15.6 & 79.2 \\
\hline
\hline
HDFLIM (C)&
35.1 & 7.3 & 35.8 & 7.1 & 31.6 & 6.1 & 35.2 & 6.8  & 71.6\\
HDFLIM (P) &
32.0 & 6.4 & 32.2 & 6.2 & 35.1 & 6.2 & 33.6 & 6.2  & 74.2 \\
\hline
\end{tabular}
\label{tab:nocaps_performance}
\end{table}

Overall, HDFLIM (C) excels in domain-specific captioning with strong semantic fidelity due to the data used during training phase, while HDFLIM (P) offers better generalization at the cost of domain-specific performance.

\paragraph{Semantic Relevance of HDFLIM predictions:} 

To show that HDFLIM predictions are semantically meaningful even when traditional metrics underestimate them, we evaluate an HDFLIM + BART pipeline. Here, HDFLIM test predictions are post-processed by a BART \cite{lewis2019_BART} model fine-tuned on HDFLIM COCO-validation outputs paired with COCO ground-truth captions, enabling it to produce benchmark-aligned captions while preserving semantic content. Table~\ref{tab:coco_BART_improvement} shows that BART post-processing substantially improves traditional metrics such as BLEU-4, METEOR, CIDEr, and SPICE. This indicates that HDFLIM’s raw predictions are already strong and generalizable, but n-gram based metrics fail to fully capture their quality. Notably, CLIP-S and RefCLIP-S scores decrease, highlighting limitations of standard evaluation metrics, consistent with prior findings \cite{hessel2021clipscore, sarto2023positive_PACS, sarto2025positive2_PACS, lee-etal-2024-fleur}. Examples of captions post-processed using BART can be found in the Appendix in Fig.~\ref{fig:example_short1} and Fig.~\ref{fig:example_short2}, where BART largely preserves the core semantic meaning of the original captions; though this is not always guaranteed while rephrasing them to improve lexical overlap with reference annotations, thereby increasing n-gram metric scores without substantive change in caption quality.

\begin{table}[h!]
\centering
\caption{Effect of BART post-processing on HDFLIM predictions for COCO. Values shown in green and red indicate improvements and degradations, respectively, for each metric.}
\begin{tabular}{lcc}
\toprule
 &  HDFLIM (C) & HDFLIM (P) \\
\midrule
BLEU@4      & \textcolor{green!60!black}{+17.6} & \textcolor{green!60!black}{+16.1} \\
METEOR        & \textcolor{green!60!black}{+6.4}  & \textcolor{green!60!black}{+8.1}  \\
CIDEr        & \textcolor{green!60!black}{+34.2} & \textcolor{green!60!black}{+43.5} \\
SPICE        & \textcolor{green!60!black}{+3.8}  & \textcolor{green!60!black}{+6.5}  \\
CLIP-S   & \textcolor{red!70!black}{-6.3}    & \textcolor{red!70!black}{-3.6}    \\
RefCLIP-S& \textcolor{red!70!black}{-3.6}    & \textcolor{red!70!black}{-1.6}    \\
\bottomrule
\end{tabular}
\label{tab:coco_BART_improvement}
\end{table}

\paragraph{Evaluating Base vs. Instruct Frozen LLM Transferability}

In real-world deployments, LLMs frequently undergo various versions of instruction tuning to enhance reasoning and safety capabilities while preserving core semantic understanding. HDFLIM learns a symbolic bridge between vision and language modalities using the Qwen2-4B Base model during training. Since this approach relies on symbolic mapping rather than deep feature alignment, a critical question emerges: \textit{Does the learned symbolic correspondence transfer when substituting the base model with its instruct variant at inference?}

\begin{figure}[h!]
    \centering
    \includegraphics[width=0.8\linewidth]{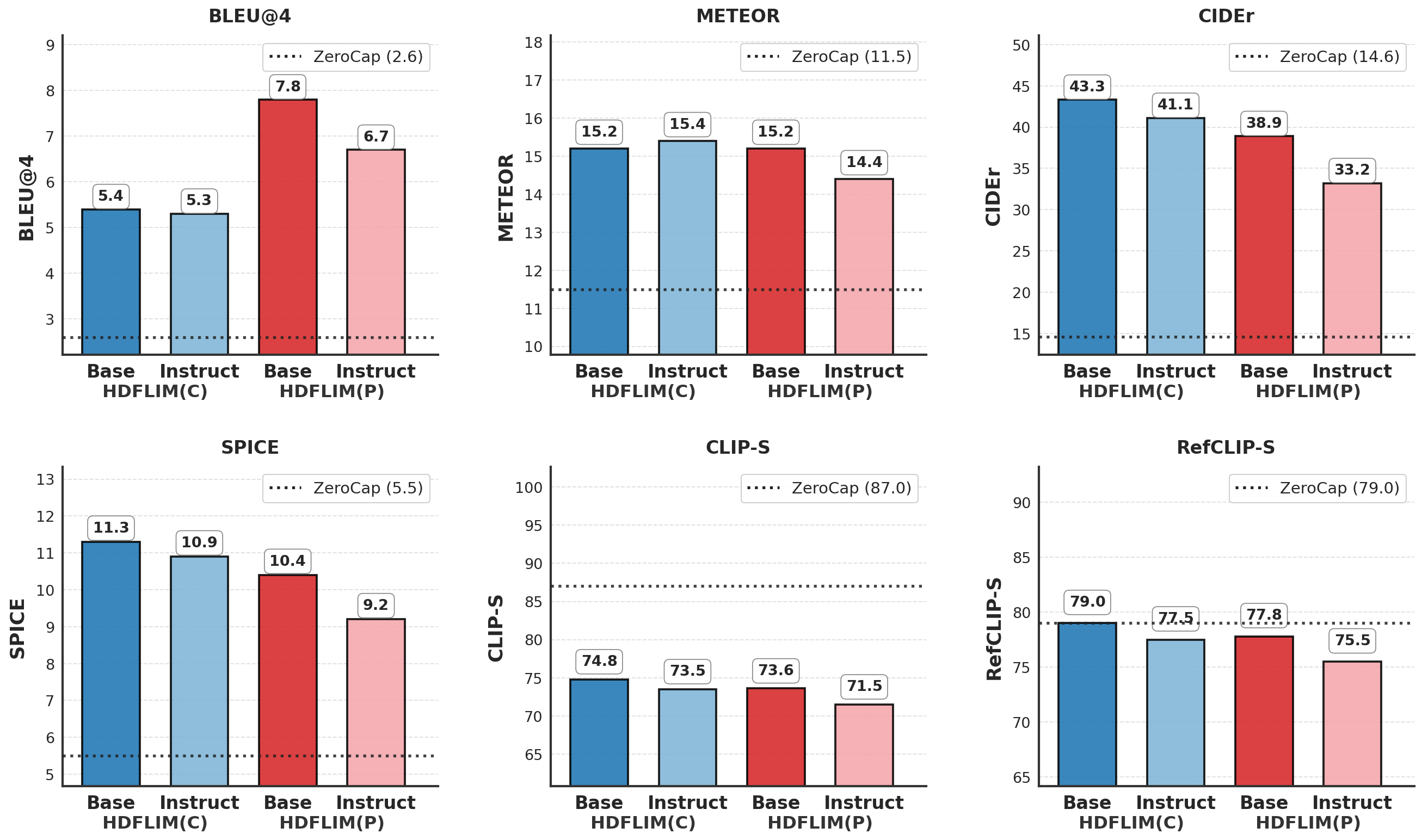}
    \caption{Captioning performance of HDFLIM variants evaluated against the ZeroCap baseline (dotted line).  HDFLIM  show transferability from Base to Instruct Qwen4B variants, with modest performance degradation, when both variants are compared to ZeroCap.}
    \label{fig:base2instruct_transferability}
\end{figure}

Instruction-tuning introduces distributional shifts that may affect decoding behavior, potentially impacting caption quality. Our experiments reveal that the learned symbolic prototypes demonstrate robust transferability across model variants. As shown in Figure~\ref{fig:base2instruct_transferability}, when replacing the base model with its instruct counterpart at inference, we observe modest performance degradation, with HDFLIM(C) experiencing minimal drops while HDFLIM(P) shows more substantial but manageable degradation.

Notably, both instruct variants maintain significant improvements over the ZeroCap baseline across traditional captioning metrics. However, we observe a consistent trade-off in CLIP-based metrics, where scores decrease when switching to instruct models. This indicates that while instruction-tuning introduces subtle semantic shifts that moderately impact the symbolic bridge's effectiveness, the learned correspondences remain sufficiently robust for practical deployment, suggesting that HDFLIM's symbolic approach achieves reasonable cross-variant generalization despite not being optimized for instruction-tuned distributions.

\paragraph{Evaluating Long-Form Captioning Performance}

\begin{figure}[h!]
    \centering
    \includegraphics[width=0.4\linewidth]{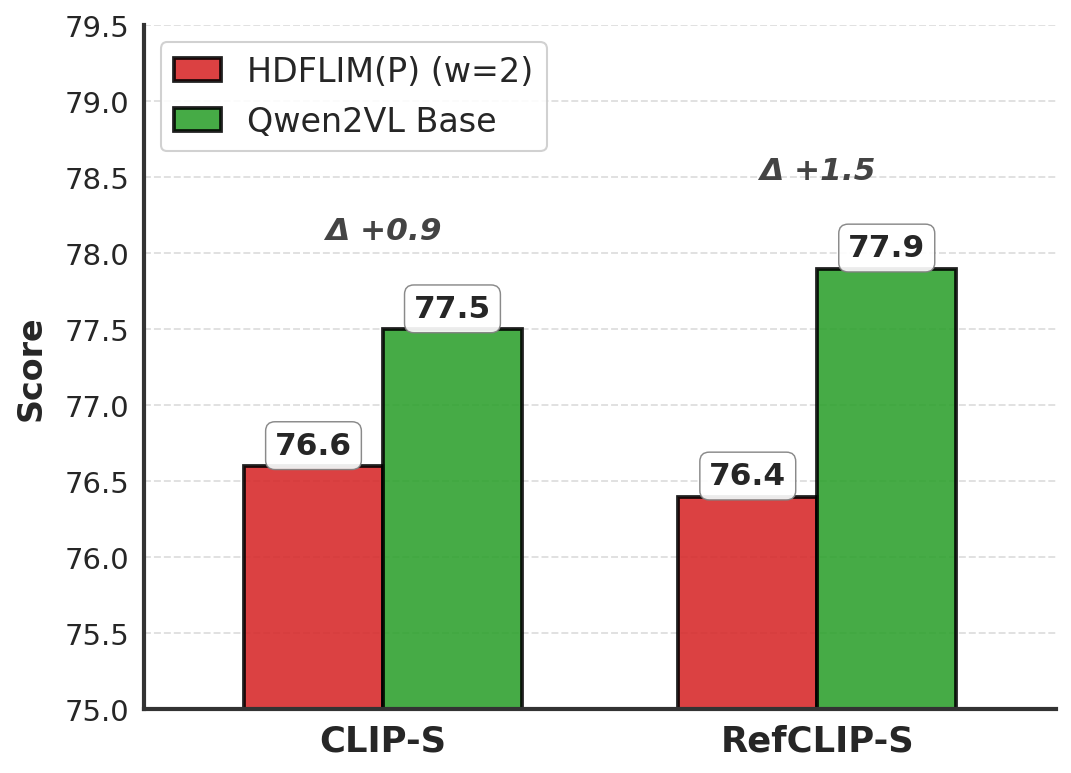}
    \caption{Reference free metric evaluation on COCO. Max number of generated tokens for HDFLIM (P) are 41 with no prompt for paraphrasing. For Qwen2VL$_{\text{Base}}$; max generated tokens are 50 with prompt: "Provide a detailed description of this image, including the main subjects, their actions, the setting, and any notable details."}
    \label{fig:long_caption_compare}
\end{figure}

We evaluate HDFLIM (P), trained on long-form image descriptions from the PixelProse dataset, to assess its capability in generating detailed, long-form image captions for COCO test split. Only CLIP-S and RefCLIP-S are used in this analysis, as traditional n-gram-based metrics (e.g., BLEU, METEOR) are not well-suited for long captions, given the significant length discrepancy between the generated descriptions and the typically concise human-annotated ground truth. We see that HDFLIM (P) achieves competitive performance; slightly below Qwen2VL$_{\text{Base}}$, but still highly comparable, as illustrated in Fig.~\ref{fig:long_caption_compare}. The performance gap is minimal, with only around a 1 to 1.5 point difference on both metrics, indicating that HDFLIM (P) effectively producescontextually appropriate long captions. Qualitative examples are provided in Fig.~\ref{fig:example_long1} and Fig.~\ref{fig:example_long2} in the Appendix.

\paragraph{Token Generation speed of HDFLIM}

\begin{figure}[h!]
    \centering
    \includegraphics[width=0.4\linewidth]{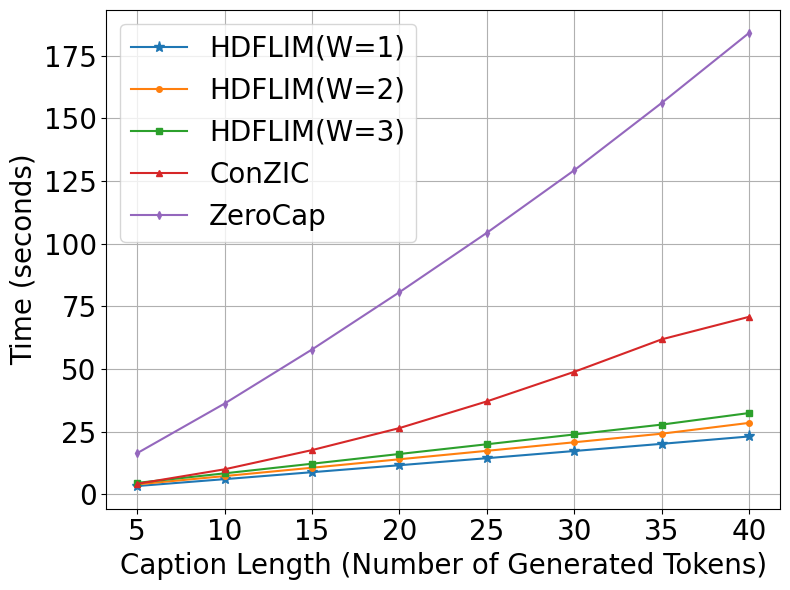}
    \caption{Inference Speed vs. Caption Length. These experiments were conducted on an NVIDIA A100 (40GB) GPU. ZeroCap and ConZIC were re-implemented for this study using the original hyperparameters reported in their respective publications to ensure fair benchmarking.}
    \label{fig:Inference_Speed}
\end{figure}

We compare the token generation speed of HDFLIM with different window size $W$ parameter, which control the search over nearby positional prototypes. Figure \ref{fig:Inference_Speed} shows a comparison against the token generation speeds of ZeroCap and ConZIC. ZeroCap requires computing gradients with respect to the CLIP loss, and ConZIC relies on repeated Gibbs sampling for polishing; as a result, both methods are significantly slower than HDFLIM, as illustrated in the figure. Their performance also further decreases as caption length increases. In contrast, HDFLIM’s token generation speed decreases only moderately as the window size $W$ increases, with minimal performance degradation. Although, one limitation of our method is that batched inference is not possible with our current implementation. 

\section{Conclusion}
In this work, we introduced HDFLIM, a hyperdimensional computing framework that establishes cross-modal alignment between frozen vision and language foundation models without parameter updates. By projecting unimodal semantic embeddings into a shared high-dimensional representational space and constructing associative memory through binding and bundling operations, HDFLIM enables direct cross-modal mapping while fully preserving pretrained representations. Unlike conventional multimodal training pipelines that rely on iterative backpropagation and large-scale fine-tuning, HDFLIM performs alignment in a single pass over the data, eliminating risks of catastrophic forgetting and substantially reducing computational overhead.

The framework’s efficiency derives from explicit high-dimensional memory construction, partial on-disk storage of associative structures, and inference-time optimization that accelerates token retrieval and generation. These properties make HDFLIM inherently suitable for scalable deployment, continual learning scenarios, and resource-constrained environments where gradient-based retraining is impractical.

Currently, HDFLIM implements a unidirectional mapping from vision to language for image captioning. Extending the same hyperdimensional alignment principles to language-to-vision generation is a straightforward architectural extension and would enable fully bidirectional multimodal reasoning within the same frozen-model paradigm.

More broadly, HDFLIM demonstrates that cross-modal alignment does not require parameter homogenization across models. Instead, semantic compatibility between independently trained foundation models can be operationalized through structured hyperdimensional interfaces that support compositional binding, robust retrieval, and modular integration. This shifts the design principle for multimodal systems from end-to-end optimization toward representational interoperability, advancing a concrete pathway toward unified world-model architectures assembled through symbolic high-dimensional alignment rather than large-scale retraining.

\section*{Acknowledgment}

This research was funded in part by grants from the National Science Foundation (2226025) to Vasant G. Honavar and the National Center for Advancing Translational Sciences of the National Institutes of Health (UL1 TR002014) (PI: Jennifer Kraschnewski, Co-I: Vasant G. Honavar); and by a Rising Researcher award ICDS\_RR26\_027475 to Abhishek Dalvi from the Penn State Institute for Computational and Data Sciences (RRID:SCR\_025154). The experiments in this work were performed on the Pennsylvania State University’s Institute for Computational and Data Sciences’ ROAR supercomputer.

%Bibliography
\bibliography{references}

\appendix
\section{Appendix}

\subsection{HDFLIM Learning, Inference and Sampling Pseudo-code}\label{appendix:Algos}

HDFLIM Learning, Inference and Sampling psuedocodes are illustrated in Algorithm \ref{algo:HDFLIM_Learn}, \ref{algo:HDFLIM_infer} and \ref{algo:HDFLIM_sample} respectively.

\begin{algorithm}
\caption{HDFLIM Learning Phase}
\label{algo:HDFLIM_Learn}
\begin{algorithmic}[1]
\setstretch{1.2}
    \State \textbf{Input:} $\mathcal{D} = \{\mathcal{I}, \mathcal{C} \}$ is the train dataset consisting of images $\mathcal{I}$ and captions $\mathcal{C}$ which are a sequence of tokens.
    % \State \textbf{Output:} The index of $key$ in $A$, or \textbf{Not Found} if absent
    \State $d_{HD} \gets 50000$
    \State \textit{// Initialize HD Prediction Matrix int32 matrix}
    \State $\textbf{HD}_{\textbf{pred}}$ $\gets$ np.memmap() to disk \Comment{Shape: $(\ell_{\max}, |\mathcal{V}|, \beta)$ }
\For{$(I, C) \in \mathcal{D}$}
    \State \textit{// Get image features using frozen vision model}
    \State $\mathbf{h}_{img} \gets \text{Encoder}_{VM}(I)$ \Comment{Shape: $(n_p, d_{I})$ }
    \State \textit{// Transform image hidden representation to HD space}
    \State $\mathbf{HD}_{img} \gets \text{transform}^{\textbf{HD}}_{img}(I)$ \Comment{Shape: $(d_{HD})$}
    \State \textit{//  Assuming $n_c \leq \ell_{\text{max}}$ is the sequence length}
    \State $C=[t_1, t_2 \ldots, t_{n_c}]$
\For{$i = 1$ \textbf{to} $(n_c-1)$} \Comment{Token 1 is a fixed prefix}
    \State \textit{// Get caption features using Frozen LLM till the $i^{th}$ token}
    \State $\mathbf{h}_{cap} \gets \text{Encoder}_{LLM}([t_1, \ldots, t_i])$ \Comment{Shape: $(i, d_C)$}
    \State \textit{//  Project last token hidden representation to HD space}
    \State $\mathbf{HD}^{(i)}_{cap} \gets \text{transform}^{\textbf{HD}}_{cap}(\mathbf{h}_{cap}^{(i)})$ 
    \Comment{Shape: $(d_{HD})$}
    \State \textit{//  Get Token ID the next token after the $i^{th}$ token}
    \State $t_{i+1} \gets C[i+1]$ 
    \State \textit{//  Combine image and $i^{th}$ token HD vectors using Binding}
    \State $\mathbf{HD}^{(i)}_{comb} \gets \mathbf{HD}_{img} \otimes \mathbf{HD}^{(i)}_{cap}$ \Comment{Shape: $(d_{HD})$}
    \State \textit{//  Add this combined HD vec to Prototype memory for $(i+1)^{th}$ token}
    \State $\textbf{HD}_{\textbf{pred}}[i+1,t_{i+1},:] = \textbf{HD}_{\textbf{pred}}[i+1,t_{i+1},:] + \mathbf{HD}^{(i)}_{comb}$
    \State \textit{//Persistent update values to disk using flushing}
    \State np.flush($\textbf{HD}_{\textbf{pred}}$)
\EndFor
\EndFor
\State \textit{//  Convert to Bipolar/Binary for Inference}
 \State $\textbf{HD}_{\textbf{pred}} = sign(\textbf{HD}_{\textbf{pred}})$
\end{algorithmic}
\end{algorithm}

\begin{algorithm}
\caption{HDFLIM Inference}
\label{algo:HDFLIM_infer}
\begin{algorithmic}[1]
\setstretch{1.3}
    \State \textbf{Input:} $I$ is the test image.
    \State \textit{// Read operation on HD Prediction Matrix bipolar/binary matrix}
    \State $\textbf{HD}_{\textbf{pred}}$ $\gets$ np.memmap() to disk \Comment{Shape: $(\ell_{\max}, |\mathcal{V}|, \beta)$ }
    \State \textit{// Get image features using frozen vision model}
    \State $\mathbf{h}_{img} \gets \text{Encoder}_{VM}(I)$ \Comment{Shape: $(n_p, d_{I})$ }
    \State \textit{// Transform image hidden representation to HD space}
    \State $\mathbf{HD}_{img} \gets \text{transform}^{\textbf{HD}}_{img}(I)$ \Comment{Shape: $(d_{HD})$}
    \State \textit{// Maintain a list for predicted tokens (Token at i=1 is a known prefix)}
    \State TokenPreds = $[t_{1}]$
    \For{$i = 1$ \textbf{to} $\ell_{\max}$} \Comment{Token 1 is a fixed prefix}
    \State \textit{// Get caption features using Frozen LLM till the $i^{th}$ token}
    \State $\mathbf{h}_{cap} \gets \text{Encoder}_{LLM}(TokenPreds)$ \Comment{Shape: $(i, d_C)$}
    \State \textit{//  Project last token hidden representation to HD space}
    \State $\mathbf{HD}^{(i)}_{cap} \gets \text{transform}^{\textbf{HD}}_{cap}(\mathbf{h}_{cap}^{(i)})$ 
    \Comment{Shape: $(d_{HD})$}
    \State \textit{//  Combine image and $i^{th}$ token HD vectors using Binding}
    \State $\mathbf{HD}^{(i)}_{comb} \gets \mathbf{HD}_{img} \otimes \mathbf{HD}^{(i)}_{cap}$ \Comment{Shape: $(d_{HD})$}
    \State \textit{//  Get logits in terms of distances for the $(i+1)^{th}$ token}
    \State $\textbf{Logits}_{\textbf{HD}} = \beta -\mathbf{d}_\mathcal{H} (\textbf{HD}_{\textbf{pred}}[i+1,:,:], \mathbf{HD}^{(i)}_{comb}$) \Comment{Shape: $(\text{vocab size})$}
    \State \textit{//  Combine HD logits and LLM logits}
    \State Get $\textbf{Logits}_{\textbf{vocab}} $ using Eq. \ref{eq:logit_deflection} \Comment{Shape: $(\text{vocab size})$}
    \State \textit{//Use a sampling algo. using these logits for predicting the token}
    \State $\hat{t}_{i+1} =$SampleToken($\textbf{Logits}_{\textbf{vocab}}$)
    \State Append $\hat{t}_{i+1}$ to TokenPreds
    \If{$\hat{t}_{i+1}$ is EOS token} 
        \State \textbf{break loop}
    \EndIf
\EndFor
\end{algorithmic}
\end{algorithm}

\begin{algorithm}
\caption{CLIP-Guided Token Sampling}
\label{algo:HDFLIM_sample}
\begin{algorithmic}[1]
\setstretch{1.3}
\State \textbf{Input:} 
    \State \quad $\textbf{logits}  \in \mathbb{R}^{|\mathcal{V}|}$: Model logits for next token ($|\mathcal{V}|$ = vocab size)
    \State \quad $\text{tokens}_{\text{pred}}$: List of previously generated token IDs
    \State \quad $\mathbf{h}_{\text{img}}$: CLIP image embedding (from frozen vision encoder)
    \State \quad $\text{clip\_weight} \in [0,1]$: Weight for CLIP guidance
    \State \quad $\text{temperature}, \text{top\_k}, \text{top\_p}, \text{repetition\_penalty}, \text{min\_candidates}$

\State \textit{// Step 1: Apply temperature scaling, repetition penalty, get top-k candidates, soft-max, and then top-p nucleus sampling}
\State $\textbf{probs}, \textbf{candidate\_tokens} \gets \text{process}(\textbf{logits})$

\State \textit{// Step 2: CLIP-guided scoring (if multiple candidates)}
\If{$\text{len}(\textbf{candidate\_tokens}) > 1$ }
    \State $\textbf{candidate\_texts} \gets [\varnothing]$
    \ForAll{$t \in \textbf{candidate\_tokens}$}
        \State $\textbf{new\_tokens} \gets \text{tokens\_pred} + [t]$
        \State $\textbf{text} \gets \text{decode}(\textbf{new\_tokens})$ \Comment{Use LLM tokenizer}
        \State $\textbf{candidate\_texts}.append(\textbf{text})$
    \EndFor

    \State \textit{// Encode candidate texts with CLIP}
    \State $\textbf{clip\_inputs} \gets \text{clip\_tokenizer}(\textbf{candidate\_texts})$
    \State $\textbf{text\_features} \gets \text{clip\_model}.encode\_text(\textbf{clip\_inputs})$

    \State \textit{// Compute CLIP similarity scores}
    \State $\textbf{clip\_scores} \gets \text{cosine\_sim}(\mathbf{h}_{\text{img}}, \textbf{text\_features})$ \Comment{Shape: (n\_candidates)}
    \State $\textbf{clip\_scores} \gets \text{softmax}(2.0 \times \textbf{clip\_scores})$ \Comment{Sharpen scores}

    \State \textit{//Combine with LM probabilities (normalized)}
    \State $\textbf{HD\_scores} \gets \textbf{probs} / \text{sum}(\textbf{probs})$
    \State $\textbf{comb\_scores} \gets \text{clip\_weight} \times \textbf{clip\_scores} + (1 - \text{clip\_weight}) \times \textbf{HD\_scores}$

    \State $\textbf{best\_idx} \gets \text{argmax}(\textbf{comb\_scores})$
\Else
    \State $\textbf{best\_idx} \gets 0$ \Comment{Fallback: pick first candidate}
\EndIf

\State \textbf{Output:} $\hat{t}_{i+1} \gets \textbf{candidate\_tokens}[\textbf{best\_idx}]$
\end{algorithmic}
\end{algorithm}

\subsection{Detailed Implementation and Model Setup}\label{appendix:implementation_details}

\paragraph{Dataset Processing}
COCO and PixelProse were downloaded and preprocessed in the WebDataset format. All images were resized such that the shortest edge was $512$ pixels, followed by a center crop to $512 \times 512$.  PixelProse dataset contains approximately 16 million image URLs with captions. We first constructed a parquet dataframe with \textit{url} and \textit{caption} columns, and then used \href{https://github.com/rom1504/img2dataset}{img2dataset} library with GNU Parallel to shard and download the dataset across a 20-node CPU cluster, saving the images in \texttt{.jpg} format. URLs with long response times (greater than 10 seconds) or those that failed after the first attempt were discarded, resulting in roughly 13 million image-caption pairs. For both COCO and PixelProse, we preprocess captions by converting them to lowercase, removing any leading article, and prepending the prefix \texttt{[This image shows ]} to each caption.

\paragraph{Hardware and Implementation Details}
All learning and inference, including advanced sampling techniques, are performed using a single NVIDIA A100-40GB GPU. Our codebase is implemented in Python and leverages NumPy (memmap) for numerical operations and PyTorch for GPU-accelerated computation of hidden representations from frozen vision models, as well as for the majority of HD operations. 

Due to the large size of the prototype matrix, the HDFLIM framework employs a combination of in-memory and on-disk computation. Specifically, partial learning and inference are performed using memory-mapped arrays on disk, which allows handling datasets larger than the available GPU or RAM memory. Additionally, bit-packing techniques are applied to speed up computation and reduce memory footprint.

\subsubsection{Partial On-Disk Learning}

Given our choice of frozen LLM: Qwen3-4B, which has a vocabulary size $|\mathcal{V}|$ of approximately 152K; our visual-linguistic prototype memory $\mathbf{HD}_{\text{pred}}$ which has data type as int32 and will have shape  $({ \ell_{max} \times 152K \times 50K})$, built during the learning phase becomes quite large. Specifically, the size of this memory is calculated as:

$$
\text{Memory Size} = \ell_{max} \times 152,000 \times 50,000 \times 4 \text{ Bytes}
$$

For a $ \ell_{max}$ of 41 (one of our configurations), this results in an array size of approximately $1.2\,\text{TB}$, and for a $ \ell_{max}$ of 21, it results in around $512\,\text{GB}$. These sizes are generally impractical to store entirely in RAM or even on a GPU due to memory limitations.

To address this, we employ $\texttt{np.memmap}$ from NumPy, which allows us to perform \textit{partial on-disk learning}. Memory mapping allows large tensors to reside on disk while being accessed as if they were in memory; the operating system transparently loads accessed pages on demand. This substantially reduces RAM requirements and enables training with datasets that exceed main memory capacity.

\subsubsection{Impact of partial learning on disk on learning speed}

Our implementation does not exactly mirror the pseudocode in Algorithm \ref{algo:HDFLIM_Learn}, as we incorporate batching during the learning phase. Batching is omitted from the pseudocode for clarity, but its inclusion is straightforward: all images are $512\times512$ and LLM representations can be padded to support batch processing without altering the algorithmic structure. Prototype matrix read/writes using \texttt{np.memmap} does not introduce a significant performance bottleneck, since all experiments were executed on the Penn State ICDS Roar cluster which comprises of an InfiniBand-backed storage system and not typical hard-disk storage system. Run time for \textbf{HDFLIM} learning phase is highly sensitive to the storage type and read/write speeds, with standard hard disks likely resulting in slower execution.

\textbf{With $\ell_{max}=21$; the learning procedure processes 50 images in 1.34 seconds, corresponding to approximately 3 hours for the full COCO dataset (approx 410,000 image–caption pairs)}. 

\textbf{For the PixelProse dataset with $\ell_{\text{max}} = 41$, the learning procedure took approximately 1.58 seconds per 50 images. The total training time was roughly five days} (excluding interrupts). However, execution was occasionally interrupted due to segmentation faults; likely caused by a large \texttt{np.flush} operation (could possibly be resolved using smaller batch size) to the disk. Despite this, since HDFLIM involves only one pass through the dataset (and instantiating the LSH matrix only once), training could be resumed from where it left off, making it more versatile for training compared to typical deep learning methods.

Note that the above run times for COCO and PixelProse do not include the binarization of the prototype memory, i.e., $\textbf{HD}_{\textbf{pred}} = \text{sign}(\textbf{HD}_{\textbf{pred}})$. This conversion involves transforming data from \texttt{int32} to \texttt{bool} and then to \texttt{packed int8} on a large \texttt{np.memmap} array. This step can take approximately 4 to 5 hours (or even less), depending on the available CPU and CPU-RAM on the system.

The HDFLIM Learning runtime can be further reduced by distributing the dataset across two or more A100 GPUs, each running an independent HDFLIM instance with separate $\mathbf{HD}_{\text{pred}}$ \texttt{int32} prototypes. The final model is obtained by summing the prototypes and applying binarization. In practice, the training time scales approximately inversely with the number of GPUs: using two GPUs reduces the runtime by about half, using three reduces it to roughly one third, and so on.

\subsubsection{Inference with Bit-Packing for Speedups}

After the learning phase, the prototype memory $\mathbf{HD}_{\text{pred}}$ is converted to a bipolar representation, which eliminates the need for using the int32 data type. Although, Python only supports 1-byte booleans; even though binary values require just one bit; we further compress the data using Bit-packing. Specifically, we group 8 values into a single byte to minimize memory usage.

As a result, the shape of $\mathbf{HD}_{\text{pred}}$ changes from:

$$
( \ell_{max} \times 152K \times 50K) \rightarrow ( \ell_{max}\times 152K \times 6250)
$$

This drastically reduces the overall memory footprint. For $ \ell_{max} = 41$, this results in approximately 41 GB of memory.

While the size still exceeds the capacity of a our single NVIDIA A100-40GB GPU setup (and even more if other steps of the algorithm are considered), this representation allows $\mathbf{HD}_{\text{pred}}$ to now fit entirely in system memory (CPU-RAM). This makes it possible to efficiently load and transfer data from CPU-RAM to the GPU, and this is further optimized using Pinned Memory (also known as "page-locked memory"). Pinned (page-locked) memory prevents the OS from paging or relocating data, enabling faster and asynchronous DMA (Direct Memory Access) transfers from RAM to the GPU.

It is also to be noted bit-packing also enhances inference performance. When computing Hamming distances during the logits computation (as in Line 18 of Algorithm \ref{algo:HDFLIM_infer}), we can directly perform bitwise operations on the bit-packed representations:

$$
\textbf{Logits}_{\textbf{HD}} = \beta - \mathbf{d}_\mathcal{H} (\textbf{HD}_{\textbf{pred}}[i+1,:,:], \mathbf{HD}^{(i)}_{comb})
$$

To compute the Hamming distance efficiently without explicitly unpacking the data, we apply a bitwise XOR operation followed by counting the number of differing bits. Since all values are bitpacked 8-bits, the resulting XOR outputs fall between 0 and 255. We use a precomputed lookup table (LUT) that maps each number between 0 and 255 to the count of its set bits, enabling fast Hamming distance computation using only integer arithmetic.

This bit-packing approach thus not only reduces memory usage but also accelerates key operations on the GPU, significantly improving inference speed. Note that we do not apply bit-packing before Line~18 of Algorithm \ref{algo:HDFLIM_infer}, as only a single HD vector is constructed prior to this step, and the packing overhead outweighs any potential performance gain.

\begin{algorithm}[H]
  \caption{Hamming Distance Computation with Lookup Table (LUT)}
  \label{alg:hamming_with_lut}
  \begin{algorithmic}[1]
    \State $\texttt{lut} = [0, 1, 1, 2, 1, 2, 2, 3, \dots, 255]$ \Comment{Precomputed bit counts}
    \State $\texttt{xor\_result} = \texttt{HD}_{\text{pred}}[i+1,:,:] \otimes \texttt{HD}^{(i)}_{comb}$ \Comment{Binding$\otimes$ is XOR operation}
    \State $\mathbf{d}_{\mathcal{H}} = sum(\texttt{lut}[xor\_result],axis=1)$ \Comment{Lookup and then sum along each row}
  \end{algorithmic}
\end{algorithm}

\subsection{Examples of HDFLIM Predictions from COCO test set} \label{appendix:example_predictions}

\begin{figure}[h!]
    \centering
    \includegraphics[width=0.5\linewidth]{sections/Figures/Example1_boysleeping.png}
    \caption{HDFLIM (P) generated Long Caption: \texttt{"Boy sleeping on a bed in his room, which is empty except for the mosquito net and some furniture. The background has an old wall with peeling paint that looks like a faded blue color."}}
    \label{fig:example_long1}
\end{figure}

\begin{figure}[h!]
    \centering
    \includegraphics[width=0.5\linewidth]{sections/Figures/example_1454.png}
    \caption{HDFLIM (P) generated Long Caption: \texttt{"Wet pavement in the rain. The people are wearing umbrellas and their faces can be seen, but they seem to have a lot of water on them or around themselves due the rain."}}
    \label{fig:example_long2}
\end{figure}

\begin{figure}[h!]
    \centering
    \includegraphics[width=0.5\linewidth]{sections/Figures/example_213.png}
    \caption{HDFLIM (C) generated Caption: \texttt{"Birds fly over the water in a cloudy day."}; transforming this prediction using BART: \texttt{"A flock of birds flying over the water."}}
    \label{fig:example_short1}
\end{figure}

\begin{figure}[h!]
    \centering
    \includegraphics[width=0.5\linewidth]{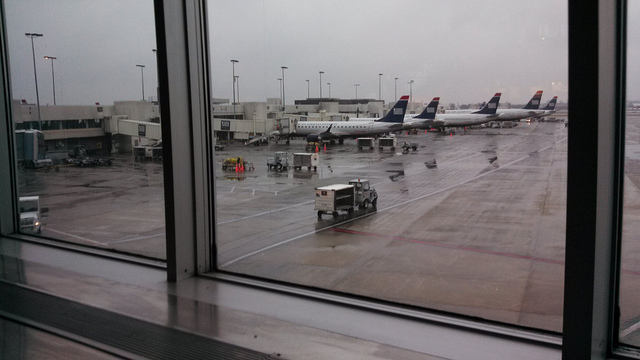}
    \caption{HDFLIM(C) generated Caption:\texttt{"Airport runway waiting planes in a terminal."}; transforming this prediction using BART: \texttt{"A plane sitting on the tarmac at an airport."}}
    \label{fig:example_short2}
\end{figure}

\subsection{Configuration and Evaluation Setup} \label{appendix:config}

At inference time, the maximum number of predicted tokens is fixed to 15. Early stopping is triggered when a full stop or \texttt{[EOS]} token is generated, consistent with the single-sentence nature of standard image captioning datasets. PixelProse, however, contains long-form and highly detailed captions. As a result, truncation at 15 tokens frequently produces incomplete sentences when using HDFLIM (P). To address this, we post-process the generated fragment by paraphrasing it with a prompt passed through the frozen language model of HDFLIM i.e Qwen3-4B, converting it into a grammatically complete caption.

PixelProse captions also tend to include highly specific details (e.g., exact product or model names), whereas benchmark datasets favor generic object descriptions. The paraphrasing prompt template is designed to normalize this specificity while preserving semantic content, ensuring stylistic compatibility with the evaluation benchmarks.

\begin{tcolorbox}[title=Prompt Template for Truncated HDFLIM (P) Predictions \\
\small{\textit{This prompt is passed to the Frozen LLM (Qwen3-4B) part of HDFLIM, i.e there are no external calls to any other models.}} ]
You are a precise image captioning assistant. You have a predicted noisy caption for the image:

Caption (predicted caption): $\{\texttt{\$CAPTION\$}\}$

Your task: Create ONE accurate, grammatically correct caption that: \\
1. Removes any contradictions or repetitions \\
2. Correct grammar \\
3. Uses natural, fluent English \\
4. Starts with "This image shows" or "The image shows" \\
5. Is ONE sentence only \\
6. Keeps all relevant objects, people, and actions mentioned \\
7. Replaces all proper nouns with generic terms and removes geolocation names \\

Output ONLY the final caption. \\

Final caption:
\end{tcolorbox}

\subsection{Extended COCO Results and Variants}\label{appendix:extended_results}

In this section, we report the numerical results corresponding to the figures presented in the main paper and include additional variants of HDFLIM + BART evaluated on the COCO test split in Tables \ref{tab:COCO_performance_appendix}, \ref{tab:COCO_performance_onQwen3diff_appendix} and \ref{tab:COCO_long_caption_performance_appendix}.

\begin{table}[h!]
\centering
\caption{Detailed Performance on Karpathy Test Split of COCO. Max number of generated tokens for HDFLIM and Qwen2VL are limited to 18. $^{\top}$values were taken from \citet{bianchi2025patchcaptionallunified_patchioner}}
\begin{tabular}{  l  | | c c c c| c c}
\hline
Model
    & B@4 & M & C & S & CLIP-S & RefCLIP-S  \\
    \hline
ZeroCap \cite{Tewel_2022_CVPR_zerocap} & 2.6 & 11.5 & 14.6 & 5.5 & 87.0 & 79.0\\

ConZIC \cite{Zeng_2023_CVPR_conzic} & 1.3 & 11.2 & 13.2 & 5.0 & 99.0 & 83.0\\
    \hline
\hline
HDFLIM (C) (w=1)     & 6.9 & 16.6 & 49.2 & 12.3 & 75.3 & 79.2 \\
HDFLIM (C) (w=1) +BART     & 24.5 & 23.0 & 83.4 & 16.1 & 69.0 & 75.6 \\
HDFLIM (C) (w=3)     & 5.4 & 15.2 & 43.3 & 11.3 & 74.8 & 79.0 \\
   HDFLIM (C) (w=3) +BART   & 24.3 & 22.9 & 82.9 & 16.2 & 69.0 & 75.4 \\
    \hline
\hline
HDFLIM (P) (w=3)     & 7.8 & 15.2 & 38.9 & 10.4 & 73.6 & 77.8 \\

HDFLIM (P) (w=3) +BART     & 23.9 & 23.3 & 82.4 & 16.9 & 70.0 & 76.2 \\
    \hline
\hline

MAGIC \cite{su2022language_MAGIC} & 12.9 & 17.2 & 48.3 & 10.9 & 74.0 & 77.0\\
ViECap$^{\top}$ \cite{fei2023transferable_ViECap} & 26.7 & 24.0 &  89.6 & 17.5  & 75.6  & -\\
\hline
\hline
CLIP-Captioner-B/32 \cite{Barraco2022TheUE_unreason_CLIPfeat} & 36.0 & 27.8 & 114.9 & 20.8 & 75.0 & 80.3\\
CLIP-Captioner-L/14 \cite{Barraco2022TheUE_unreason_CLIPfeat} & 38.7 & 29.3 & 126.0 & 22.5 & 75.4 & 81.1\\
Qwen2VL$_{\text{Base}}$ \cite{wang2024qwen2vlenhancingvisionlanguagemodels}     & 16.9 & 26.0 & 47.1 & 20.3 & 81.0 & 81.9\\
Qwen2VL$_{\text{FT}}$ \cite{wang2024qwen2vlenhancingvisionlanguagemodels}       & 40.0& 30.7 & 137.5  & 24.2 & 78.6 & 84.0\\
\hline
\end{tabular}
\label{tab:COCO_performance_appendix}
\end{table}

\begin{table}[h!]
\centering
\caption{Performance of transferability of learned symbolic prototype across various Finetuned Version for Karpathy Test Split of COCO when Qwen3-4B Base model is replaced with Qwen3-4B-Instruct for Inference. Across all results, HDFLIM Learning was done on only Qwen3-4B base model. ZeroCap and ConZIC are kept as baselines. (w=3)}
\begin{tabular}{  l  | | c c c c| c c}
\hline
Model
    & B@4 & M & C & S & CLIP-S & RefCLIP-S  \\
    \hline
\multicolumn{7}{c}{}  \\[-8pt]
\multicolumn{7}{c}{\textit{No Training; Test time gradient compute or Sampling}}  \\[4pt]
\hline
ZeroCap \cite{Tewel_2022_CVPR_zerocap} & 2.6 & 11.5 & 14.6 & 5.5 & 87.0 & 79.0\\

ConZIC \cite{Zeng_2023_CVPR_conzic} & 1.3 & 11.2 & 13.2 & 5.0 & 99.0 & 83.0\\
    \hline
\multicolumn{7}{c}{}  \\[-8pt]
\multicolumn{7}{c}{\textit{HDFLIM with Learning phase on COCO dataset}}  \\[4pt]
\hline
HDFLIM$_{\text{Qwen3-Base}}$     & 5.4 & 15.2 & 43.3 & 11.3 & 74.8 & 79.0 \\
HDFLIM$_{\text{Qwen3-Instruct}}$     & 5.3 & 15.4 & 41.1 & 10.9 & 73.5 & 77.5 \\
\hline
\multicolumn{7}{c}{}  \\[-8pt]
\multicolumn{7}{c}{\textit{HDFLIM with Learning phase on PixelProse dataset}}  \\[4pt]
\hline
HDFLIM$_{\text{Qwen3-Base}}$     & 7.8 & 15.2 & 38.9 & 10.4 & 73.6 & 77.8 \\
HDFLIM$_{\text{Qwen3-Instruct}}$     & 6.7 & 14.4 & 33.2 & 9.2 & 71.5 & 75.5 \\
\hline
\end{tabular}
\label{tab:COCO_performance_onQwen3diff_appendix}
\end{table}

\begin{table}[h!]
\centering
\caption{Performance on Karpathy Test Split of COCO. Max number of generated tokens for HDFLIM are 41 and Qwen2VL is 50 with prompt: "Provide a detailed description of this image, including the main subjects, their actions, the setting, and any notable details."}
\begin{tabular}{l | c c}
\hline
Model & CLIP-S & RefCLIP-S \\
\hline
\hline
HDFLIM (P) (w=3) & 76.6 & 76.4 \\
\hline
\hline
Qwen2VL$_{\text{Base}}$ \cite{wang2024qwen2vlenhancingvisionlanguagemodels} & 77.5 & 77.9 \\
\hline
\end{tabular}
\label{tab:COCO_long_caption_performance_appendix}
\end{table}

\subsection{Cross-Domain Evaluation on the Flickr Test Split}
We additionally evaluate our method on the \texttt{Flickr30k Karpathy} split \cite{young_Flickr30k, karpathy2015deep}. This experiment measures cross-domain performance, where models trained on the COCO training set are evaluated on the Flickr test set.

\begin{table}[h!]
\centering
\caption{Cross-Domain Performance on Flickr30K Karpathy Test split; Trained/Learned on COCO; and Tested on Flickr30K Karpathy Test split}
\begin{tabular}{  l  | | c c c c| c c}
\hline
Model
    & B@4 & M & C & S & CLIP-S & RefCLIP-S  \\
    \hline
HDFLIM (C) (w=3)    & 6.3 & 12.8 & 26.5 & 8.7 & 73.1 & 75.8 \\
    \hline
\hline
MAGIC \cite{su2022language_MAGIC} & 6.2 & 12.2 & 17.5 & 5.9 & 68.6 & - \\
CapDec \cite{nukrai-etal-2022-text_CapDec}  & 17.3 & 18.6 & 35.7 & - & 73.7 & - \\
ViECap \cite{fei2023transferable_ViECap} & 17.4 & 18.0 & 38.4 & 11.2 & 76.1  & - \\
\hline
\hline
Qwen2VL$_{\text{FT}}$ \cite{wang2024qwen2vlenhancingvisionlanguagemodels} & 32.1 & 25.5 & 88.3 & 18.5 & 78.7 & 82.3 \\
\hline
\end{tabular}
\label{tab:Flickr_performance_appendix}
\end{table}

% \begin{table}[h!]
% \centering
% \caption{Long caption Performance on Karpathy Test Split of COCO. Max number of generated tokens for HDFLIM are 41 and Qwen2VL for  is tokens=30 new=50 with prompt "Provide a detailed description of this image, including the main subjects, their actions, the setting, and any notable details."}
% \begin{tabular}{  l  | | c c| c c}
% \hline
% Model
%     & M & S & CLIP-S & RefCLIP-S  \\
%     \hline
% \multicolumn{5}{c}{}  \\[-8pt]
% \multicolumn{5}{c}{\textit{HDFLIM with Learning phase on PixelProse dataset}}  \\[4pt]
% \hline
% HDFLIM (P) (w=2)     & 17.2 & 12.3 & 76.6 & 76.4 \\
% \hline
% \hline
% Qwen2VL$_{\text{Base}}$ \cite{wang2024qwen2vlenhancingvisionlanguagemodels}    & 20.1 & 15.9 & 77.5 & 77.9 \\
% \hline
% \end{tabular}
% \label{tab:COCO_long_caption_performance}
% \end{table}

\subsection{Analysis of CLIP Weight and Window Parameters}

In this section, we analyze the performance of HDFLIM to two key hyperparameters: the CLIP weighting factor used during sampling and the window size $W$. Tables \ref{tab:COCO_performance_withclipweightting} and \ref{tab:COCO_performance_withwindow} report performance variations across captioning metrics as these parameters are adjusted. This study highlights the trade-offs between semantic alignment and caption quality and demonstrates the robustness of HDFLIM across a range of parameter settings. The results also show that CLIP-weighted sampling provides a substantial performance boost; however, HDFLIM remains effective even without it, albeit at comparatively lower performance.

\begin{table}[h!]
\centering
\caption{Performance of HDFLIM (C) with w=3 with varying Clip-weight parameter during sampling for Test Split of COCO dataset}
\begin{tabular}{  l  | | c c c c| c c}
\hline
CLIP-Weight
    & B@4 & M & C & S & CLIP-S & RefCLIP-S  \\
    \hline

0.00    & 7.0 & 13.6 & 26.8 & 8.4 & 59.2 & 65.7 \\
0.25    & 7.7 & 15.2 & 42.5 & 11.0 & 70.0 & 75.4 \\
0.50    & 5.4 & 15.2 & 43.3 & 11.3 & 74.8 & 79.0 \\
0.75    & 2.1 & 13.8 & 35.3 & 9.5 & 77.1 & 80.0 \\
1.00    & 0.2 & 12.7 & 25.0 & 8.0 & 77.5 & 78.4 \\

\hline
\end{tabular}
\label{tab:COCO_performance_withclipweightting}
\end{table}

\begin{table}[h!]
\centering
\caption{Performance of HDFLIM (C) with varying window parameter during sampling for Karpathy Test Split of COCO dataset}
\begin{tabular}{  l  | | c c c c| c c}
\hline
W
    & B@4 & M & C & S & CLIP-S & RefCLIP-S  \\
    \hline

1    & 6.9 & 16.6 & 49.2 & 12.3 & 75.3 & 79.2 \\

2    & 6.2 & 15.8 & 46.1 & 11.7 & 75.0 & 79.0 \\

3    & 5.4 & 15.2 & 43.3 & 11.3 & 74.8 & 79.0 \\

\hline
\end{tabular}
\label{tab:COCO_performance_withwindow}
\end{table}

\end{document}